\documentclass{article}

\usepackage[final]{corl_2020} 
\usepackage{times}

\usepackage[numbers]{natbib}
\usepackage{multicol}
\usepackage[normalem]{ulem}
\usepackage[ruled, linesnumbered]{algorithm2e}  

\usepackage{amsfonts}
\usepackage{amsopn}
\usepackage{amsmath}
\usepackage{amssymb}
\usepackage{mathtools}
\usepackage{bbm}
\usepackage{color}
\usepackage{placeins}
\usepackage{wrapfig}
\usepackage{float}
\usepackage{nicefrac}

\newcommand{\E}[2]{\operatorname{\mathbb{E}}_{#1}\left[#2\right]}

\newcommand{\policy}{\pi}

\newcommand{\state}{\mathbf{s}}
\newcommand{\sz}{{\state_0}}

\newcommand{\st}{{\state_t}}

\newcommand{\stp}{{\state_{t+1}}}

\newcommand{\action}{\mathbf{a}}

\newcommand{\at}{{\action_t}}

\newcommand{\qparams}{\theta}
\newcommand{\pparams}{\phi}
\newcommand{\sparams}{\psi}

\newcommand{\aref}[1]{\hyperref[#1]{Appendix~\ref*{#1}}}

\def\alignautorefname~#1\null{%
  (#1)\null
}\def\equationautorefname~#1\null{%
  Equation~#1\null
}
\title{Learning to Walk in the Real World \\ with Minimal Human Effort}

\author{
    Sehoon Ha$^{12*}$, Peng Xu$^{2}$, Zhenyu Tan$^{2}$, Sergey Levine$^{2,3}$, and Jie Tan$^{2}$ \\
    $^{1}$Georgia Institute of Technology $^{2}$ Robotics at Google \\ 
    $^{3}$Berkeley Artificial Intelligence Research, University of California, Berkeley \\
    Email: \mbox{sehoonha@gatech.edu, \{pengxu,tanzheny,jietan\}@google.com},svlevine@berkeley.edu\\
    $^{*}$The research was conducted when the author was at Google.
}

\begin{document}
\maketitle


\begin{abstract}
    Reliable and stable locomotion has been one of the most fundamental challenges for legged robots. Deep reinforcement learning (deep RL) has emerged as a promising method for developing such control policies autonomously. In this paper, we develop a system for learning legged locomotion policies with deep RL in the real world with minimal human effort. The key difficulties for on-robot learning systems are automatic data collection and safety. We overcome these two challenges by developing a multi-task learning procedure and a safety-constrained RL framework. We tested our system on the task of learning to walk on three different terrains: flat ground, a soft mattress, and a doormat with crevices. Our system can automatically and efficiently learn locomotion skills on a Minitaur robot with little human intervention. The supplemental video can be found at: \url{https://youtu.be/cwyiq6dCgOc}.
\end{abstract}



\section{Introduction}

Reliable and stable locomotion has been one of the most fundamental challenges in the field of robotics.
Traditional hand-engineered controllers often require expertise and manual effort to design. While this can be effective for a small range of environments, it is hard to scale to the large variety of situations that the robot may encounter in the real world. In contrast, deep reinforcement learning (deep RL) can learn control policies automatically, without any prior knowledge about the robot or the environment. In principle, each time the robot walks on a new terrain, the same learning process can be applied to automatically acquire an optimal controller for that environment.

However, despite the recent successes of deep reinforcement learning, these algorithms are often exclusively evaluated in simulation. Building fast and accurate simulations to model the robot and the rich environments that the robot may be operating in are extremely difficult. For this reason, we aim to develop a deep RL system that can learn to walk autonomously \emph{in the real world}. This learning system needs to be scalable so that many robots can interact with and collect data simultaneously in a large variety of real-world environments, and eventually learn from this data. There are many challenges to designing such a system. In addition to finding a stable and efficient deep RL algorithm, the key challenge is to reduce human effort during the training process. Only when little human supervision is needed, can the learning system be truly scalable.




In this paper, we focus on solving two bottlenecks of reducing human efforts for robot learning: automation and safety. During training, the robot needs to automatically and safely retry the locomotion task hundreds or even thousands of times. This requires the robot staying within the training area, minimizing the number of dangerous falls, and automating the resets between trials. We accomplish all these via a multi-task learning procedure and a safety-constrained learner.
By simultaneously learning to walk in different directions, the robot stays within the training area. By automatically balancing between reward and safety, the robot falls dramatically less frequently.

Our main contribution is an autonomous real-world reinforcement learning system for robotic locomotion, 
which allows a quadrupedal robot to learn multiple locomotion skills on a variety of surfaces with minimal human intervention. 
We test our system to learn locomotion skills on flat ground, a soft mattress and a doormat with crevices.
Our system can learn to walk on these terrains in just a few hours, with minimal human effort, and acquire distinct and effective gaits for each terrain. In contrast to the prior work \cite{haarnoja2018learning}, in which over a hundred manual resets are required in the simple case of walking on flat ground, our system requires \emph{zero} manual resets in this case.
We also show that our system can train four policies simultaneously (walking forward, backward and turning left and right), which form a complete skill-set for navigation and can be composed into an interactive directional walking controller at test time.
\section{Related Work} 
\label{sec:related}

Control of legged robots is typically decomposed into a modular design, such as state estimation, foot-step planning, trajectory optimization, and model-predictive control.
For instance, researchers have demonstrated agile locomotion with quadrupedal robots using a state-machine~\cite{BledtPKCWK18}, impulse scaling~\cite{park2017high}, and model predictive control~\cite{katz2019mini}.
The ANYmal robot \cite{hutter2016anymal} plans footsteps based on the inverted pendulum model \cite{raibert1986legged}, which is further modulated by a vision component \cite{tsounis2019deepgait}.
Similarly, bipedal robots can be controlled by online trajectory optimization~\cite{apgar2018fast} or whole-body control~\cite{kim2016stabilizing}.
These approaches have been used for many locomotion tasks, from stable walking to highly dynamic running, but often require considerable prior knowledge of the target robotic platform and the task.
Instead, we aim to develop an end-to-end on-robot training system that automatically learns locomotion skills from real-world experience, which requires no prior knowledge about the dynamics of the robot and the environment.
Note that our approach is different from repertoire-based adaptation~\cite{cully2015robots,chatzilygeroudis2018reset} that relies on the pre-obtained gait library.

Recently, deep reinforcement learning has drawn attention as a general framework for acquiring control policies.
It has been successful for finding effective policies for various robotic applications, including autonomous driving~\cite{amini2019variational}, navigation~\cite{chiang2018learning}, and manipulation~\cite{kalashnikov2018qt}.
Deep RL also has been used to learn locomotion control policies \cite{heess2017emergence}, mostly in simulated environments.
Despite its effectiveness, one of the biggest challenges is to transfer the trained policies to the real world, which often incurs significant performance degradation due to the \emph{sim-to-real} gap.
Although researchers have developed principled approaches for mitigating the sim-to-real issue, including system identification \cite{Hwangboeaau5872}, domain randomization \cite{tan2018sim,chebotar2019closing}, and meta-learning \cite{finn2017model,yu2019learning}, it remains an open problem.

Researchers have investigated applying RL to real robotic systems directly, which is intrinsically free from the sim-to-real gap~\cite{tedrake2005learning,morimoto2007improving}.
The approach of learning on real robots has achieved state-of-the-art performance on manipulation and grasping tasks, by collecting a large amount of interaction data on real robots \cite{levine2016end,kalashnikov2018qt,zeng2019tossingbot}.
However, applying the same method to underactuated legged robots is challenging. One major challenge is the need to reset the robot to the proper initial states after each episode of data collection, for hundreds or even thousands of roll-outs.
The pioneer work of \citet{kohl2004policy} developed an automated environment to learn locomotion, in which a reasonable starting policy is needed to avoid mechanical failures from unstable gaits.
Researchers tackled the resetting issue further by using statically-stable robots \cite{li2019learning}, or developing external resetting devices for lightweight robots, such as a simple 1-DoF system~\cite{ha2018automated} or an articulated robotic arm~\cite{luck2017lab}. Otherwise, the learning process requires many manual resets between roll-outs~\cite{haarnoja2018learning,choi2019trajectory,yang2019data}, which limits the scalability of the learning system.

Another challenge is to ensure the safety of the robot during the entire learning process.
Safety in RL can be formulated as a constrained Markov Decision Process (cMDP)~\cite{altman1999constrained}, which can be solved by the Lagrangian relaxation procedure~\cite{bertsekas1997nonlinear}.
\citet{achiam2017constrained} discussed a theoretical foundation of cMDP problems, which guarantees to improve rewards while the safety constraints are satisfied.
Many researchers have proposed extensions of the existing deep RL algorithms to address safety, such as learning an additional safety layer that projects raw actions to a safe feasibility set~\cite{alshiekh2018safe,chow2019lyapunov},
learning the boundary of the safe states with a classifier~\cite{lipton2016combating,tigas2019robust},
expanding the identified safe region progressively~\cite{berkenkamp2017safe},
or training a reset policy alongside with a task policy~\cite{eysenbach2018leave}. 

In this paper, we focus on developing an autonomous and safe learning system for legged robots, which can learn locomotion policies with minimal human intervention. Closest to our work is the prior paper by \citet{haarnoja2018learning}, which also uses soft actor-critic (SAC) to train walking policies in the real world. In contrast to that work, our focus is on eliminating the need for human intervention, while the prior method requires a person to intervene hundreds of times during training. We further demonstrate that we can learn locomotion on challenging terrains, and simultaneously learn multiple policies, which would be too tedious or even infeasible in the prior work \cite{haarnoja2018learning} due to the number of human interventions needed.
\section{Background: Reinforcement Learning}
\label{sec:preliminary}
We formulate the task of learning to walk in the setting of reinforcement learning~\cite{sutton1998reinforcement}.
The problem is represented as a Markov Decision Process (MDP), which is defined by the state space $\mathcal{S}$, action space $\mathcal{A}$, stochastic transition function $p(\stp | \st,\at)$, reward function $r(\state_t, \action_t)$, and the distribution of initial states $\state_0 \sim p(\state_0)$.
By executing a policy $\pi(\action_t | \st)\in\Pi$, we can generate a trajectory of state and actions $\tau = (\state_0,\action_0, \state_1, \action_1, \ldots)$.
We denote the trajectory distribution induced by $\policy$ by $\rho_\pi(\tau)=p(\sz)\prod_t\pi_t(\at|\st)p(\stp|\st,\at)$. Our goal is to find the optimal policy that maximizes the sum of expected returns:
\begin{equation}
    J(\pi) = \E{\tau\sim \rho_\policy}{\sum_{t = 0}^{T} r(\st, \at)}.
    \label{eq:rl_objective}
\end{equation}

\section{Automated Learning in the Real World} 
\label{sec:method}

\begin{figure}
    \centering
    \setlength{\tabcolsep}{1pt}
    \renewcommand{\arraystretch}{0.7}
    \begin{tabular}{c c c}
    \includegraphics[width=0.24\textwidth]{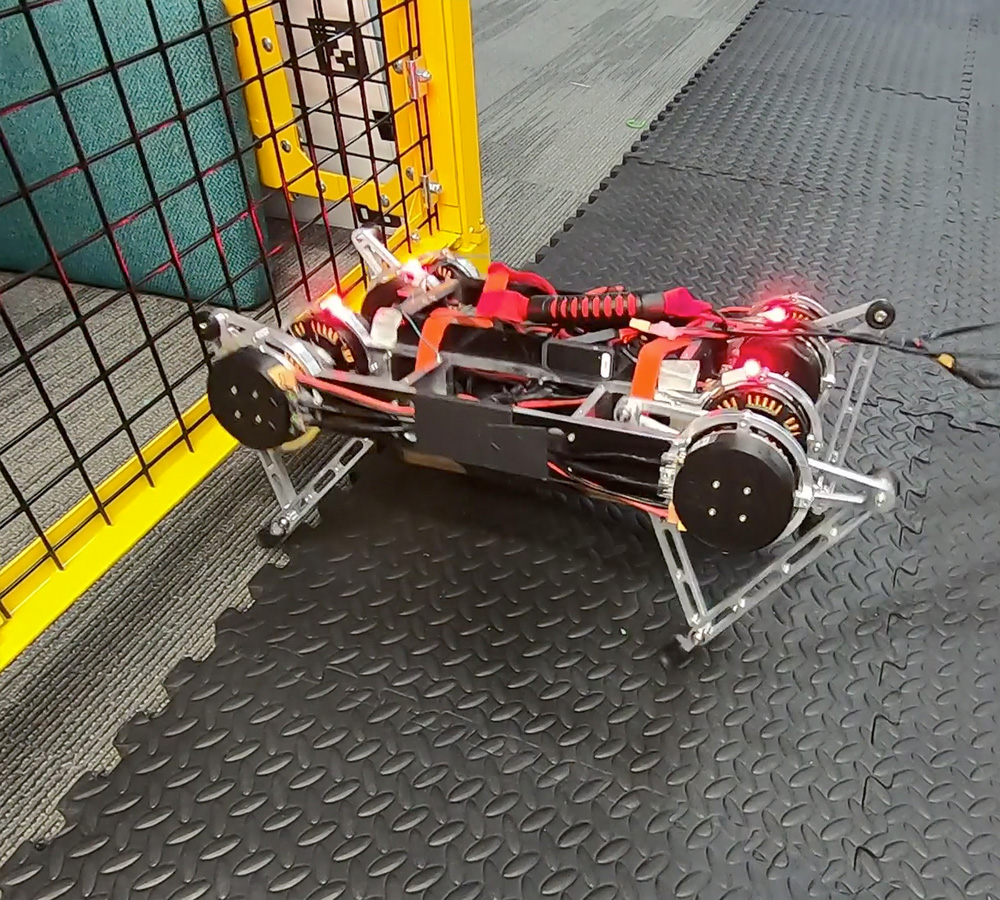} &
    \includegraphics[width=0.24\textwidth]{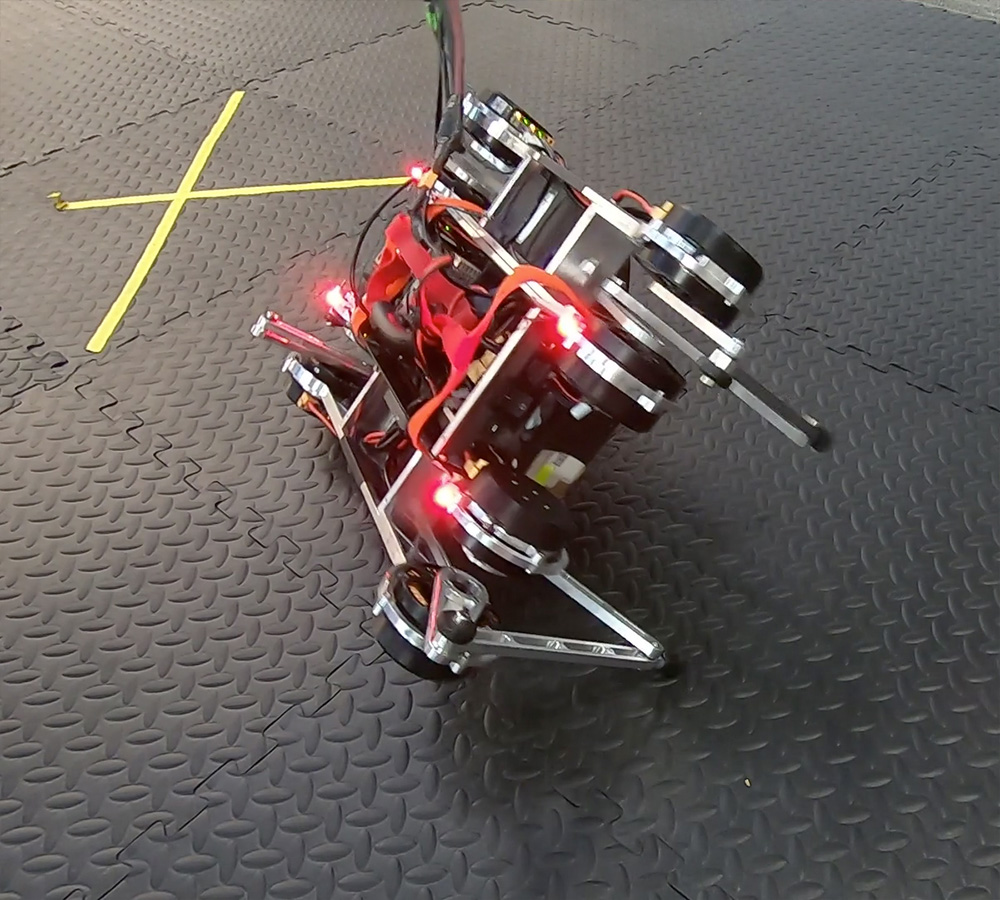}
    &
    \includegraphics[width=0.48\textwidth]{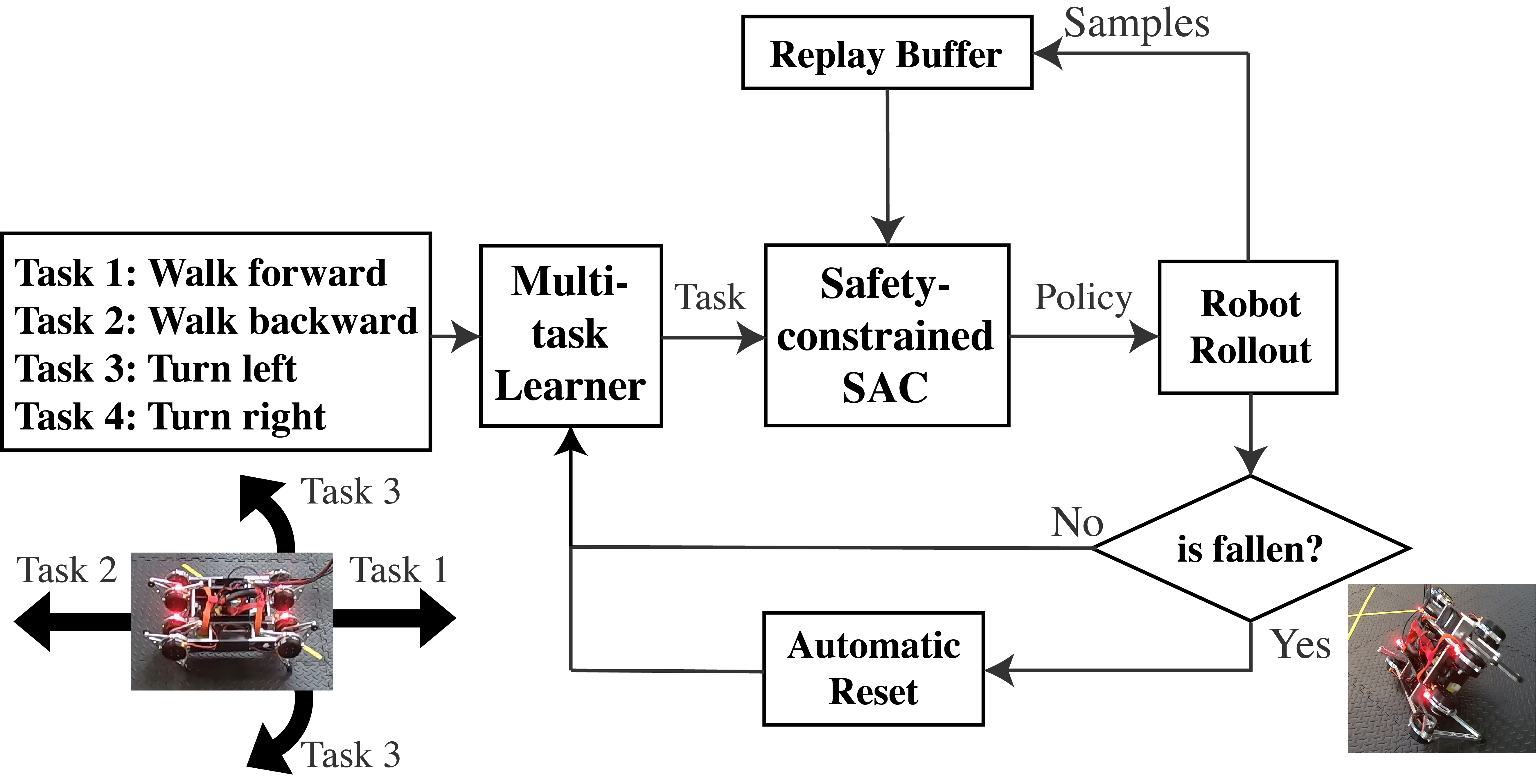}
    \\
    \end{tabular}
    \caption{\small An autonomous learning system for real legged robots requires to address two challenges: (\textbf{Left}) a robot leaving the workspace and (\textbf{Middle}) a robot falling to the ground.
    (\textbf{Right}) Overview of our learning system. We solve main automation challenges, leaving the workspace, falling, and resetting, by multi-task learning and a safety-constrained RL algorithm .
    }
	\label{fig:failure_and_overview}
\end{figure}

Our goal is to develop an automated system for learning locomotion controllers in the real world with minimal human intervention (Figure~\ref{fig:failure_and_overview}). Aside from incorporating a stable and efficient deep RL algorithm, our system needs to address the following challenges. 
First, the robot must remain within the training area (\emph{workspace)}, which is difficult if the system only learns a single policy that walks in one direction. We utilize a multi-task learning framework, which simultaneously learns \emph{multiple} tasks for walking and turning in different directions. This multi-task learner selects the task to learn according to the relative position of the robot in the workspace. For example, if the robot is about the leave the workspace, the selected task-to-learn would be walking backward. Using this simple state machine scheduler, the robot can remain within the workspace during the entire training process. 
Second, the system needs to minimize the number of falls because falling can result in substantially longer training times due to the overhead of experiment reset after each fall. Even worse, the robot can get damaged after repeated falls. We augment the Soft Actor-Critic (SAC) \cite{haarnoja2018sac} formulation with a safety constraint that limits the roll and the pitch of the robot's torso. Solving this cMDP greatly reduces the number of falls during training. 
Our system, combining these components, dramatically reduces the number of human interventions in most of the training runs.

\subsection{Automated Unattended Learning via Multi-Task RL}
\label{sec:multitask}



One main reason of manual intervention for learning locomotion is the need to move the robot back to the initial position after each episode. Otherwise, the robot would leave the workspace within a few rollouts. 
We develop a multi-task learning method with a state-machine scheduler that generates an interleaved schedule of multi-directional locomotion tasks. In our formulation, a task is defined by the desired direction of walking with respect to its initial position and orientation at the beginning of each roll-out.
More specifically, the task reward $r$ is parameterized by a three dimensional task vector $\mathbf{w}^i = [w^i_1, w^i_2, w^i_3]^T$:
\begin{equation}
r_\mathbf{w^i}(\state, \action) = [w^i_1, w^i_2]^T \cdot \mathbf{R}_0^{-1}(\mathbf{x}_t - \mathbf{x}_{t-1})
+ w^i_3 (\theta_t - \theta_{t-1}) - 0.001 |\ddot{\mathbf{a}}|^2,
\label{eq:reward}
\end{equation}
where $\mathbf{R}_0$ is the rotation matrix of the robot's torso at the beginning of the episode, $\mathbf{x}_t$ and $\theta_t$ are the position and yaw angle of the torso in the horizontal plane at time $t$, and $\ddot{\mathbf{a}}$ measures smoothness of actions, which is the desired motor acceleration in our case.
This task vector $\mathbf{w}^i$ defines the desired direction of walking. For example, walking forward is $[1, 0, 0]^T$ and turning left is $[0, 0, 0.5]^T$.
Note that the tasks are locally defined and invariant to the selection of the starting position and orientation of each episode.
For each task $\mathbf{w}^i$ in the predefined task set \mbox{$\mathcal{W} = \{\mathbf{w}^1, \cdots, \mathbf{w}^n\}$}, we train a separate policy $\pi^i(a|s)$.

\begin{wrapfigure}{r}{0.4\textwidth}
  \vspace{-2mm}
  \begin{center}
    \includegraphics[width=0.38\textwidth]{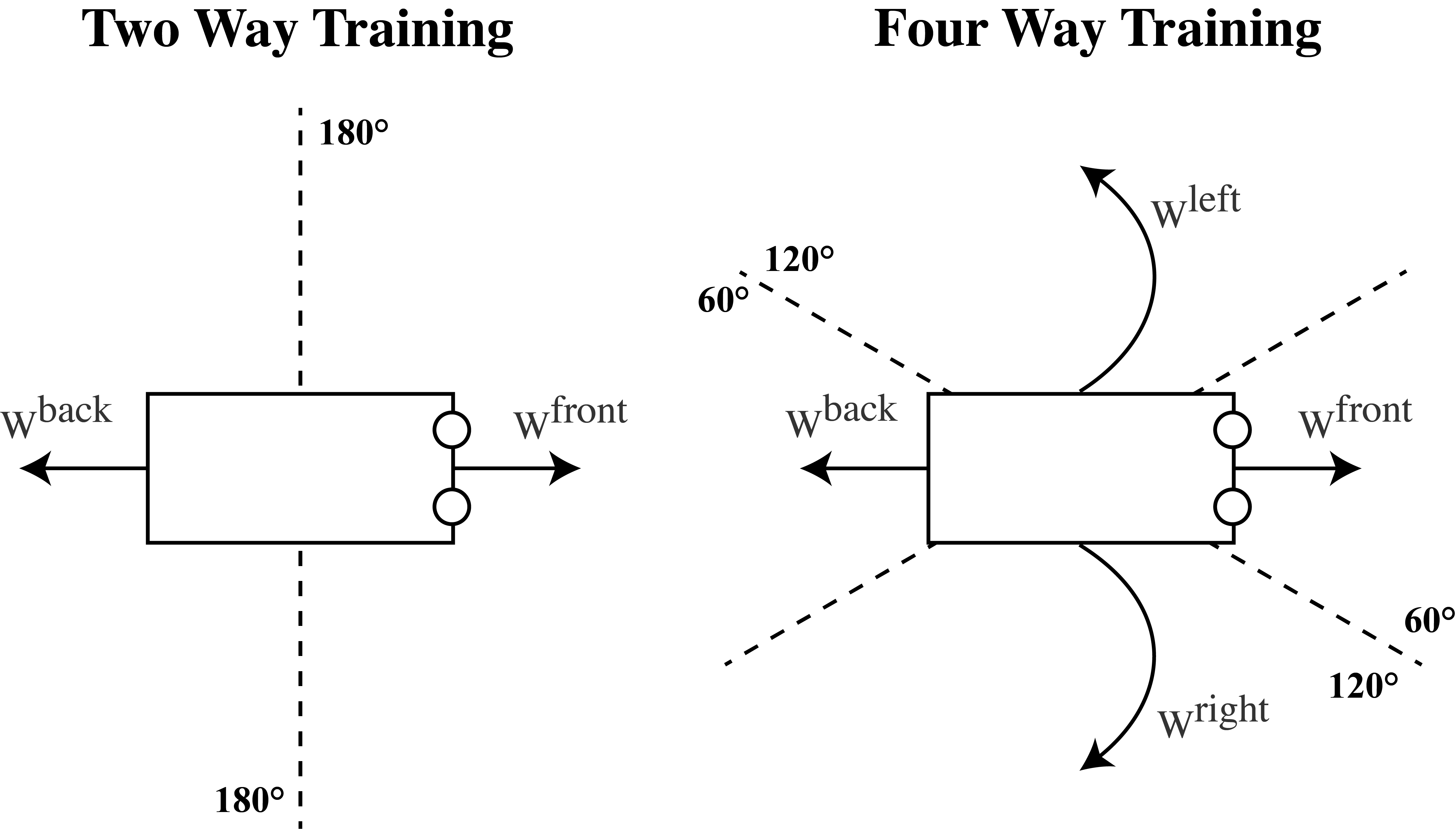} 
  \end{center}
\end{wrapfigure}
At the beginning of the episode, the scheduler determines the next task-to-learn based on the relative position of the center of the workspace in the robot's coordinate. 
In effect, our scheduler selects the task in which the desired walking direction is pointing towards the center. This is done by dividing the workspace in the robot's coordinate with fixed angles and selecting the task where the center is located in its corresponding subdivision.
For example if we have two tasks, forward and backward walking, the scheduler will select the forward task if the workspace center is in front of the robot, and select the backward task in the other case. Note that a simple round-robin scheduler will not work because the tasks may not be learned at the same rate.

Our multi-task learning method is based on two assumptions:
First, we assume that even a partially trained policy still can move the robot in the desired direction by a small amount most of the time. In practice, we find this to be true. While the initial policy cannot move the robot far away from the center, as the policy improves, the robot quickly begins to move in the desired direction, even if it does so slowly and unreliably at first.
Usually after $10$ to $20$ roll-outs, the robot starts to move at least $1$~cm, by pushing its torso in the desired direction.
The second assumption is that, for each task in the set $\mathcal{W}$, there is a counter-task that moves in the opposite direction. For example, walking forward versus backward or turning left versus right.
Therefore, if one policy drives the robot to the boundary of the workspace, its counter-policy can bring it back.
In our experience, both assumptions hold for most scenarios, unless the robot accidentally gets stuck at the corners of the workspace.

We train a policy for each task with a separate instance of learning, without sharing actors, critics, or replay buffer. We made this design decision because we did not achieve clear performance gain when we experimented with weights and data sharing, probably because the definition of tasks is discrete and the experience from one task may not be helpful for other tasks. 

Although multi-task learning reduces the number of out-of-workspace failures by scheduling proper task-to-learn between episodes, the robot can still occasionally leave the workspace if it travels a long distance in one episode. We prevent this failure by triggering early termination when the robot is near and moving towards the boundary. In contrast to falling, this early termination requires special treatments for return calculation. Since the robot does not fall and can continue executing the task if it is not near the boundary, we take future rewards into consideration when we compute the target values of Q functions.

\subsection{RL with Safety Constraints}
\label{sec:safetyrl}

Repeated falls may not only damage the robot, but also significantly slow down training, since the robot must stand up after it falls.
In fact, each reset takes more than $12$ seconds (more details in the Appendix), which is longer than the duration of an entire rollout.
To mitigate this issue, we formulate a constrained MDP to find the optimal policy that maximizes the sum of rewards while satisfying the given safety constraints $f_s$:
\begin{equation}
    \max_{\pi\in\Pi} \E{\tau\sim\rho_\pi}{\sum_{t=0}^T r(\st,\at)}
    \text{ s.t. } \E{(\st,\at)\sim\rho_\pi}{f_s(\st,\at)} \geq 0,\ \forall t.
    \label{eq:constrained_rl}
\end{equation}
In our implementation, we design the safety constraints to prevent falls that can easily damage the servo motors:
\begin{equation}
    f_s(\st, \at) = \min(\hat{p} - |p_t|,\hat{r} - |r_t|)
    \label{eq:safety}
\end{equation}
where $p$ and $r$ are the pitch and roll angles of the robot's torso, and $\hat{p}$ and $\hat{r}$ are the maximum allowable tilt, where they are set to $\pi/12$ and $\pi/6$ for all experiments.

We can rewrite the constrained optimization by introducing a Lagrangian multiplier $\lambda$:
\begin{equation}
    \mathcal{L}(\policy, \lambda) = \mathbb{E}_{\tau\sim\rho_\pi}\bigg[\sum_{t=0}^T r(\st,\at)  
    + \lambda f_s(\st,\at)
    \bigg].
    \label{eq:lagrangian}
\end{equation}
We optimize this objective using the dual gradient descent method~\cite{boyd2004convex},
which alternates between the optimization of the policy $\policy$ and the Lagrangian multiplier $\lambda$.
First, we train both Q functions for the regular reward $Q_{\qparams}$ and the safety term $S_{\sparams}$, which are parameterized by $\qparams$ and $\sparams$ respectively, using the formulation similar to \citet{haarnoja2018learning}.
Then we can obtain the following actor loss:
\begin{align}
     \E{\st\sim\mathcal{D}, \at\sim\policy_\pparams}{- Q_{\qparams}(\st, \at) - \lambda S_{\sparams}(\st, \at)},
\label{eq:policy_objective}
\end{align}
where $\mathcal{D}$ is the replay buffer.
Finally, we can learn the Lagrangian multiplier $\lambda$ by minimizing the loss $J(\lambda)$:
\begin{align}
\E{\st\sim \mathcal{D}, \at\sim \policy_{\pparams}}{ \lambda f_s(\at, \st)}.
\label{eq:alpha_objective}
\end{align}
The important steps of the algorithm is summarized in Algorithm~\ref{alg:learning}.

\begin{algorithm}
\caption{Automated Learning with Safety Constraint}
\label{alg:learning}
Initialize replay buffer $D^{1 \cdots n}=\{\}$, function approximators $\qparams$, $\sparams$, policy $\pi$, Lagrangian multiplier $\lambda$\\
\For{each episode}{
    Sample a task $\mathbf{w}^k$ from the multi-task scheduler \tcp{Section \ref{sec:multitask}}
	\For{each environment step}{
	$\at \sim \policy^k(\at|\st)$ \\
	$\stp \sim p(\stp| \st, \at)$ \\
	$D^k\leftarrow D^k\cup \{\st, \at, \stp, r_\mathbf{w}^k(\st, \at)\}$\\
	Gradient update on $Q^k_{\qparams}$ and $\policy^k$   \tcp{Standard SAC steps}
	Gradient update on $S^k_{\sparams}$ and $\lambda^k$   \tcp{Safety constraint steps (Section \ref{sec:safetyrl})}
	}
}
\end{algorithm}

\section{Experiments}
\label{sec:experiments}

We design experiments to validate that our proposed system can learn locomotion controllers in the real world with minimal human intervention. 
Particularly, we aim to answer the following questions.
\begin{enumerate}
    \item Can our system learn locomotion policies in the real world with minimal human effort?
    \item Can our system learn multiple policies simultaneously?
    \item Can our safety-constrained RL learner reduce the number of falls?
\end{enumerate}

\subsection{Learning on Flat Terrain}
\begin{figure}
    \vspace{-0.3cm}
    \centering
    \setlength{\tabcolsep}{1pt}
    \renewcommand{\arraystretch}{0.7}
    \begin{tabular}{c c c c}
    \includegraphics[width=0.245\textwidth]{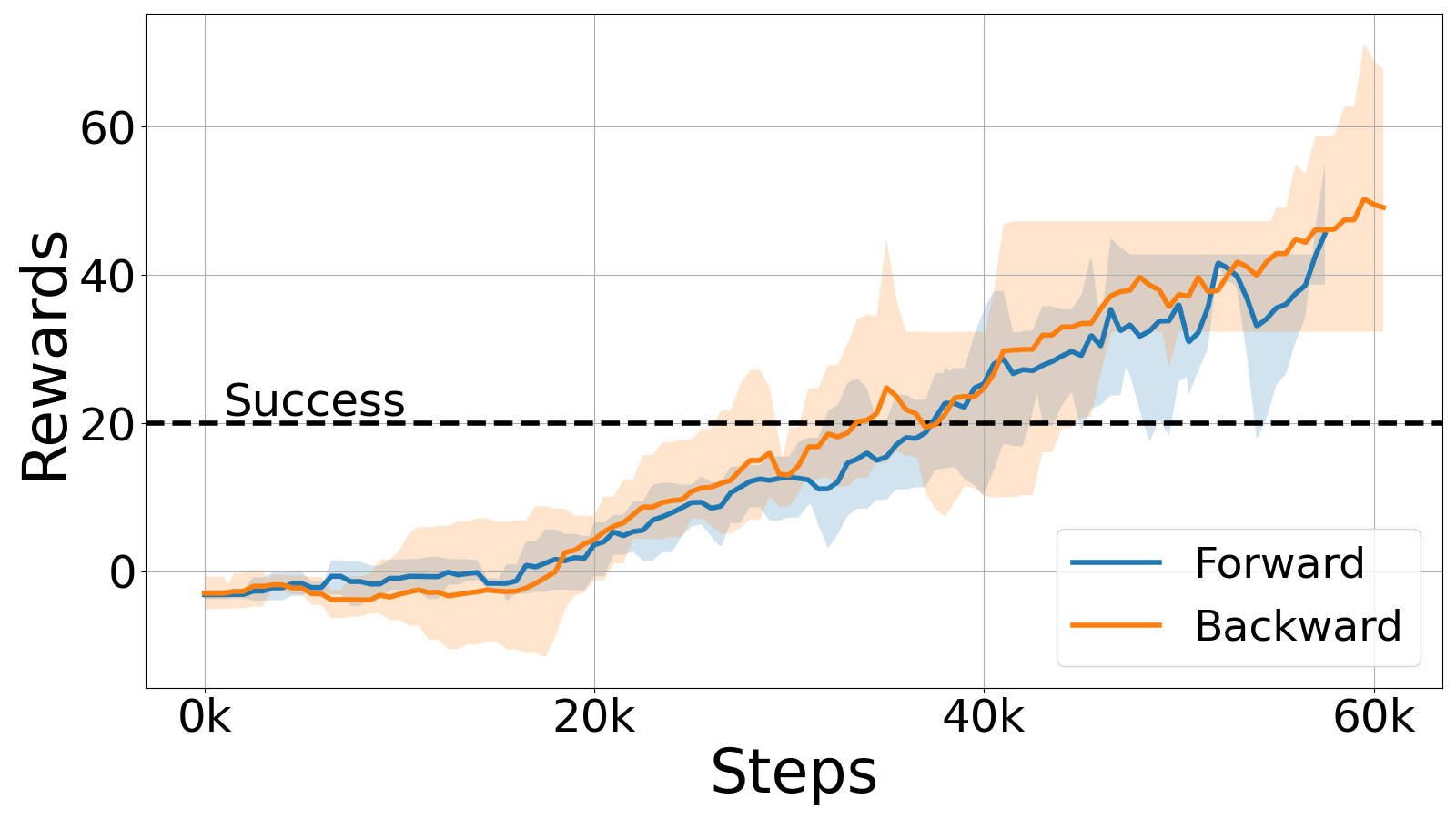} &
    \includegraphics[width=0.245\textwidth]{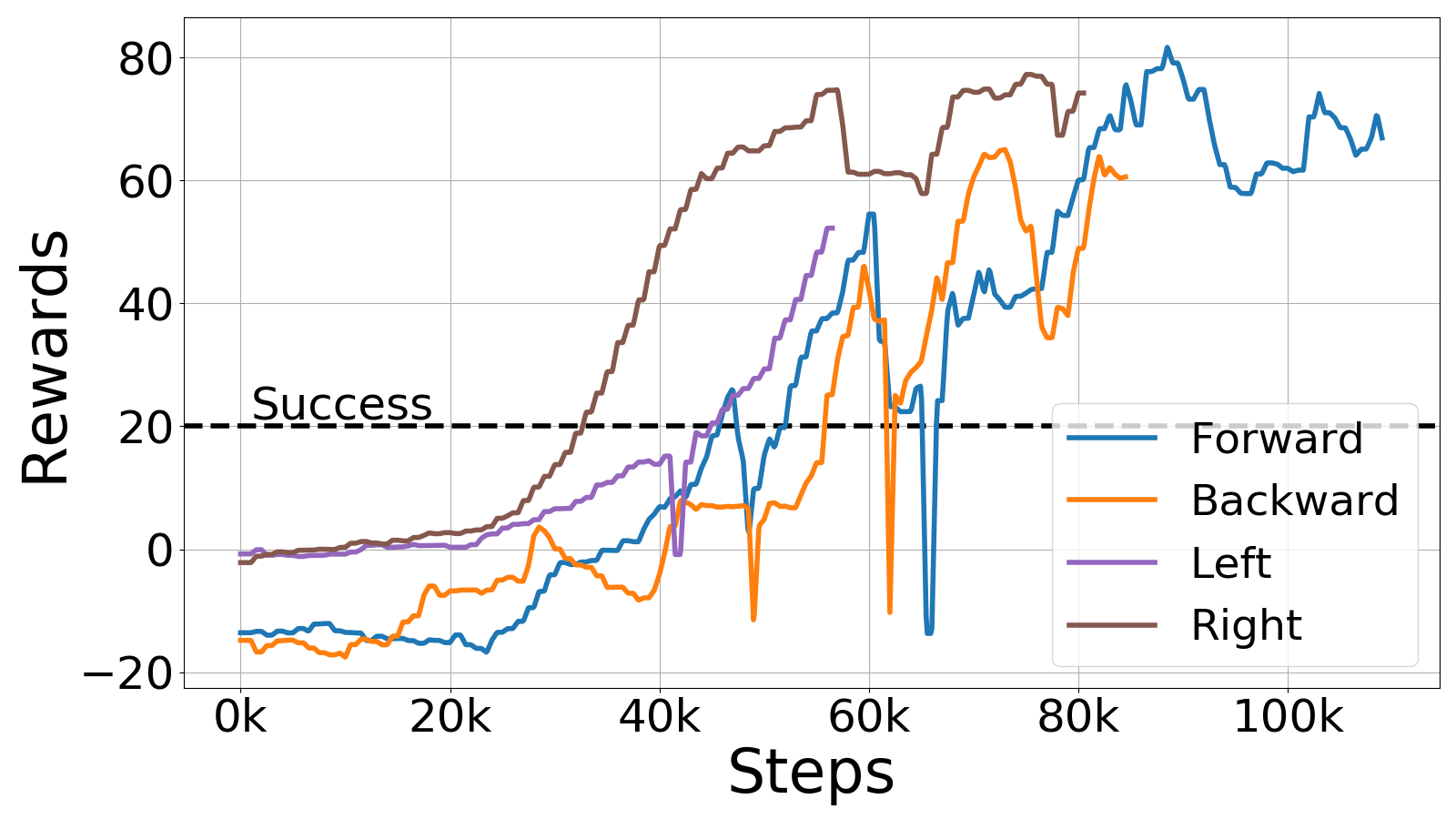} &
    \includegraphics[width=0.245\textwidth]{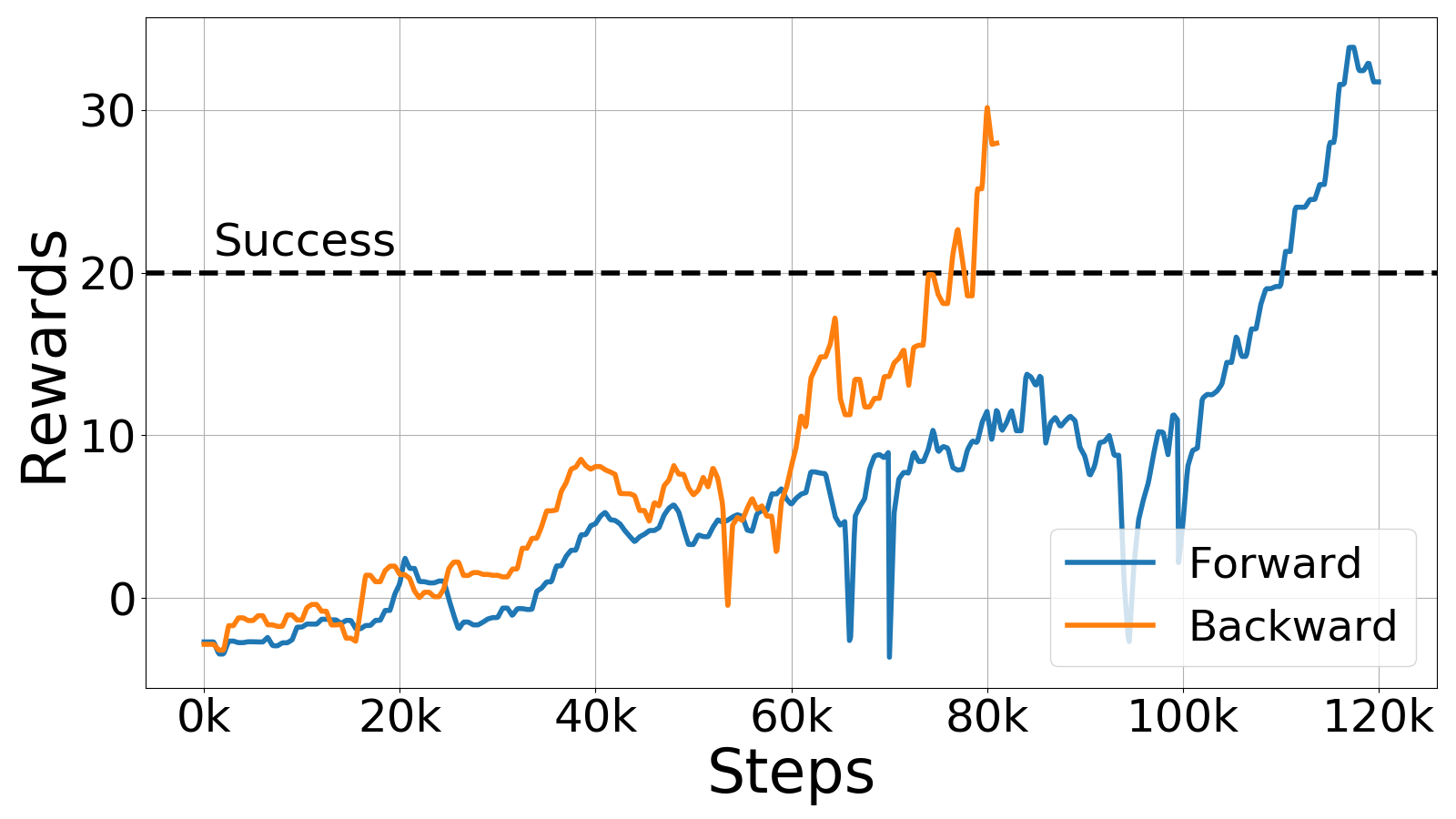} &
    \includegraphics[width=0.245\textwidth]{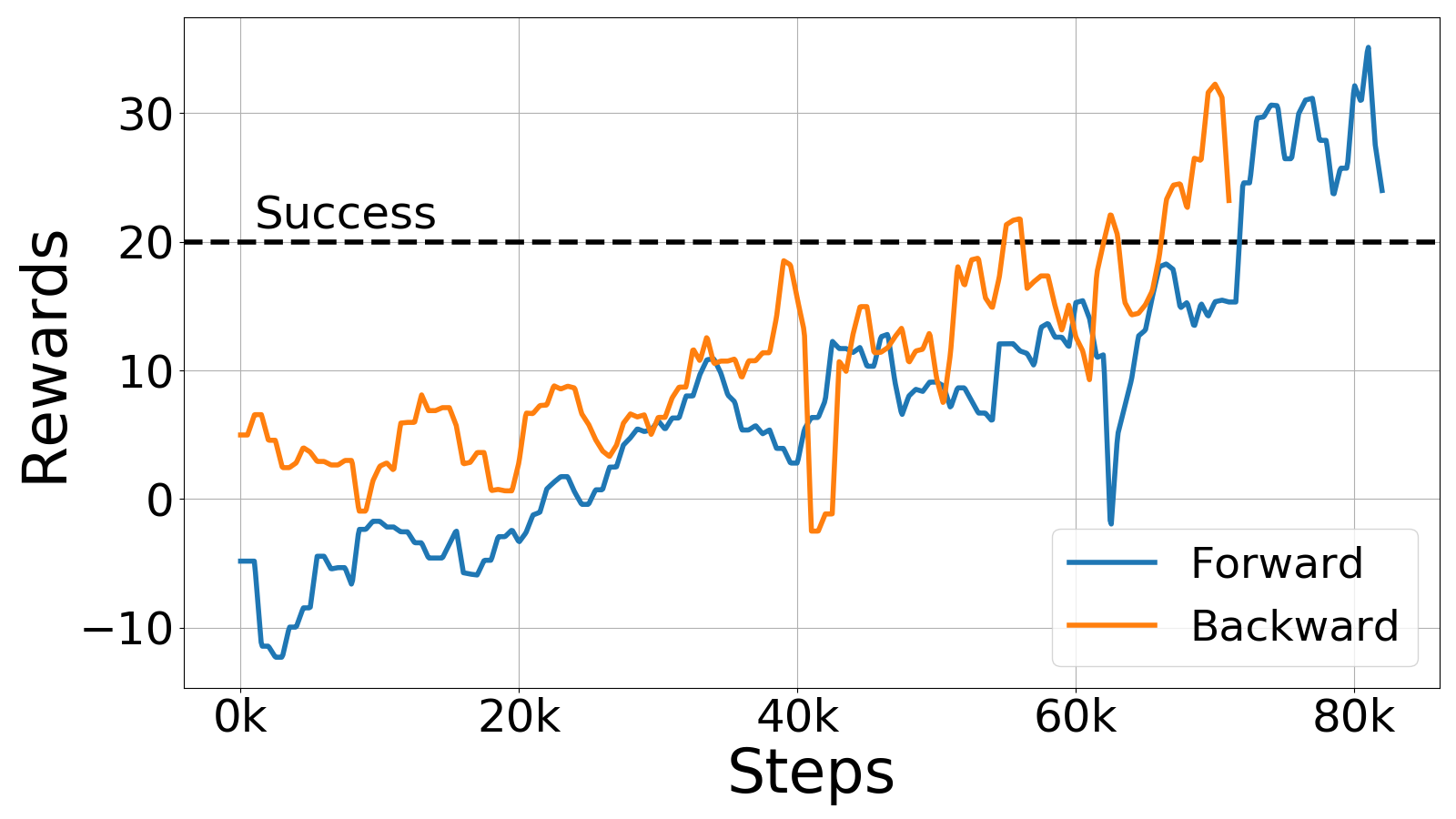} \\    
    \end{tabular}
    \caption{\small Learning curves of hardware experiments.
    \textbf{(1st and 2nd}) Learning curves on flat terrain with two tasks and four tasks. In both cases, our system learns successful multi-directional walking policies (reward $\geq 20.0$) with minimal human intervention. The shaded area in the first plot represents the min and max of three trials.
    (\textbf{3rd and 4th}) Learning curves on the soft mattress and the doormat with crevices.
    On both challenging surfaces, our framework learns successfully (reward $\geq 20.0$).
    }
	\label{fig:learning_curves}
\end{figure}

In the first experiment, we test our system using a Minitaur, a quadruped robot (Figure~\ref{fig:failure_and_overview}) \cite{kenneally2016design}, to learn forward and backward walking in a $5 \times 2 m^2$ training area with flat ground. Please refer to the Appendix for the details of the experiment setup. In two of three repeated training runs, our method requires zero human interventions. The third run only needs two manual resets when the robot was stuck at the corner. 
In contrast, the prior work~\cite{haarnoja2018learning}, which did not have the multi-task learner, required hundreds of human interventions during a single training run. 
In addition, our system requires far less data:
while the prior work~\cite{haarnoja2018learning} needed 2 hours or 160k steps to learn a single policy, our system trains two policies (forward and backward walking) in 1.5 hours ($\sim$60k steps per policy). Figure~\ref{fig:learning_curves} (1st) shows the learning curve for both tasks. 
This greatly-improved efficiency is due to our system updating the policy every step of the real-world execution, which turns out to be much more sample efficient than the episodic, asynchronous update scheme of \citet{haarnoja2018learning}.

\begin{figure*}
    \centering
    \setlength{\tabcolsep}{1pt}
    \renewcommand{\arraystretch}{0.7}
    \begin{tabular}{c c c c c}
    \includegraphics[width=0.195\textwidth]{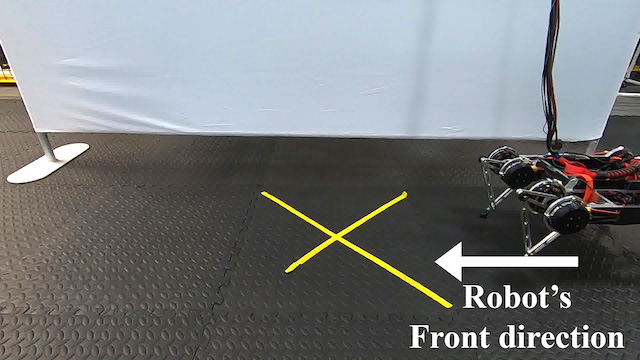} &
    \includegraphics[width=0.195\textwidth]{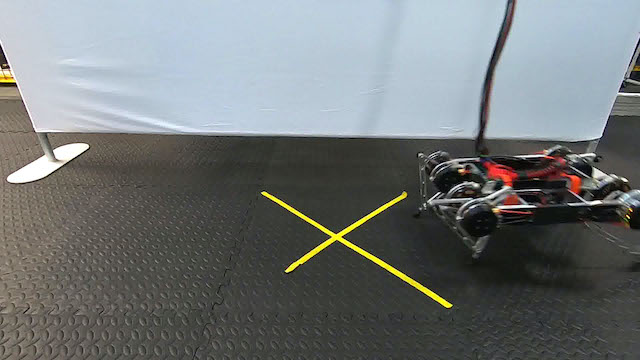} &
    \includegraphics[width=0.195\textwidth]{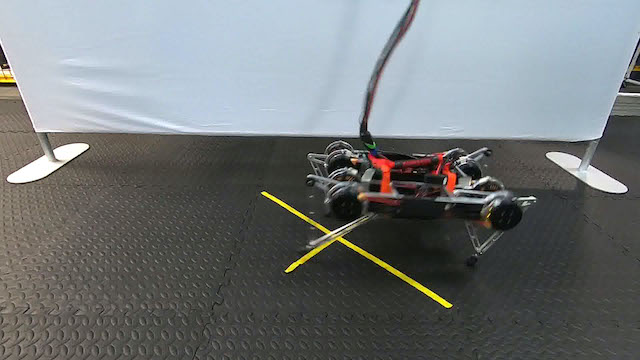} &
    \includegraphics[width=0.195\textwidth]{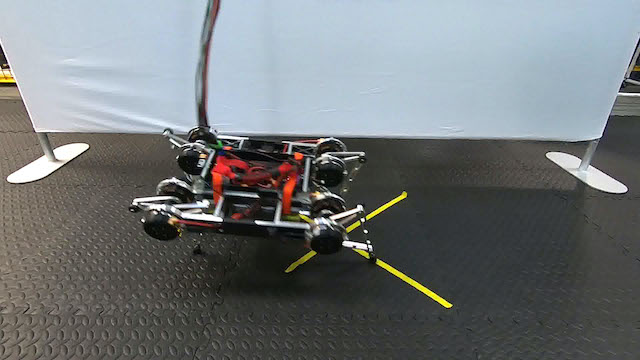} &
    \includegraphics[width=0.195\textwidth]{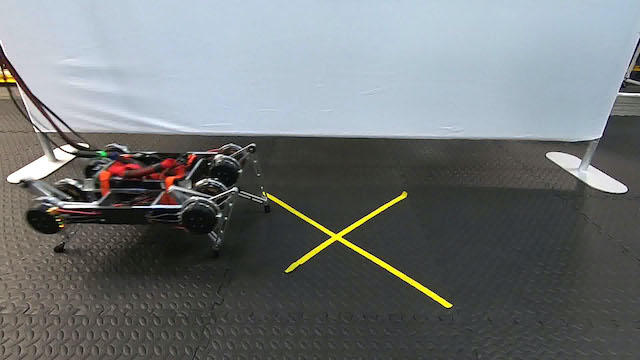} \\
    \includegraphics[width=0.195\textwidth]{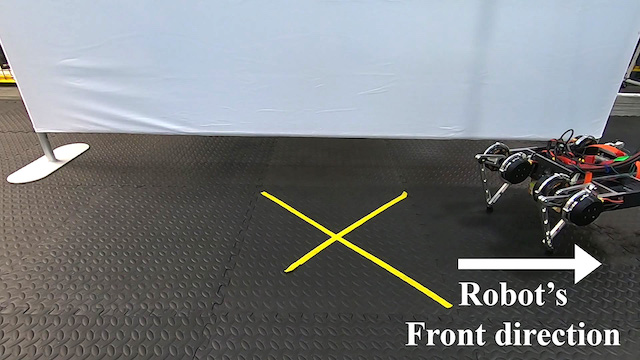} &
    \includegraphics[width=0.195\textwidth]{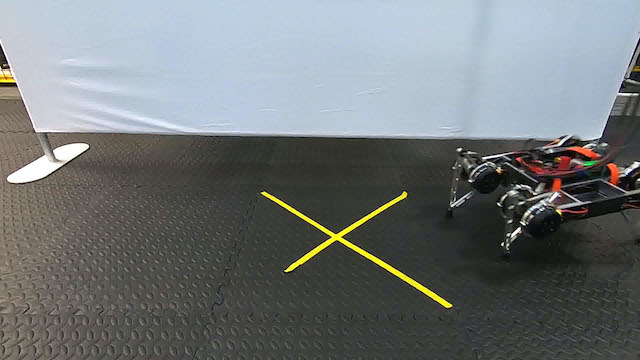} &
    \includegraphics[width=0.195\textwidth]{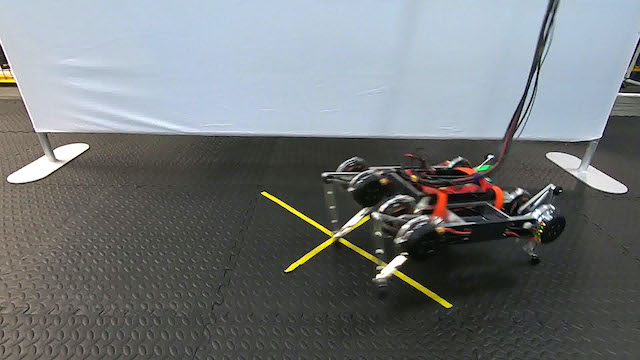} &
    \includegraphics[width=0.195\textwidth]{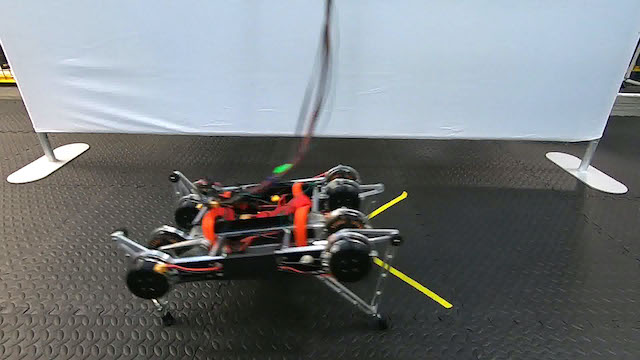} &
    \includegraphics[width=0.195\textwidth]{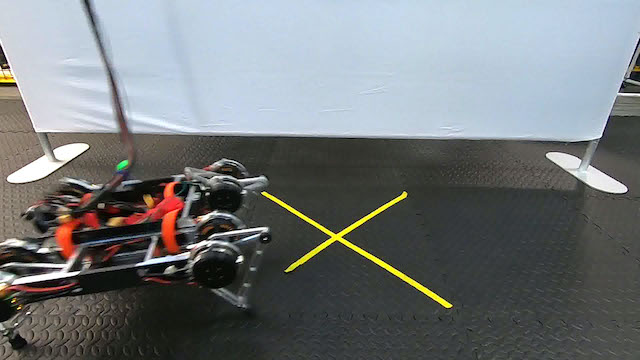} \\
    \end{tabular}
    
    \caption{Learned policies on the flat terrain. (\textbf{Top}) The forward policy  moves the legs on the same side synchronously, which resembles a pacing gait. 
    (\textbf{Bottom}) The backward gait is similar to a bounding gait by having left-right symmetric motions. 
    }
	\label{fig:flat_motion}	
\end{figure*}

We observe that the robot's asymmetric leg structure leads to different gaits for walking forward and backward. 
Forward walking requires higher ground clearance to the forward-facing feet. 
In contrast, the robot can drag the feet when it walks backward.
As a consequence, the learned forward walk resembles a regular pacing gait with high ground clearance, and the backward walk resembles a high-frequency bounding gait.
Please refer to Figure~\ref{fig:flat_motion} and the supplemental video for more detailed illustrations.

\begin{wrapfigure}{r}{0.4\textwidth}
  \vspace{-2mm}
  \begin{center}
    \includegraphics[width=0.38\textwidth]{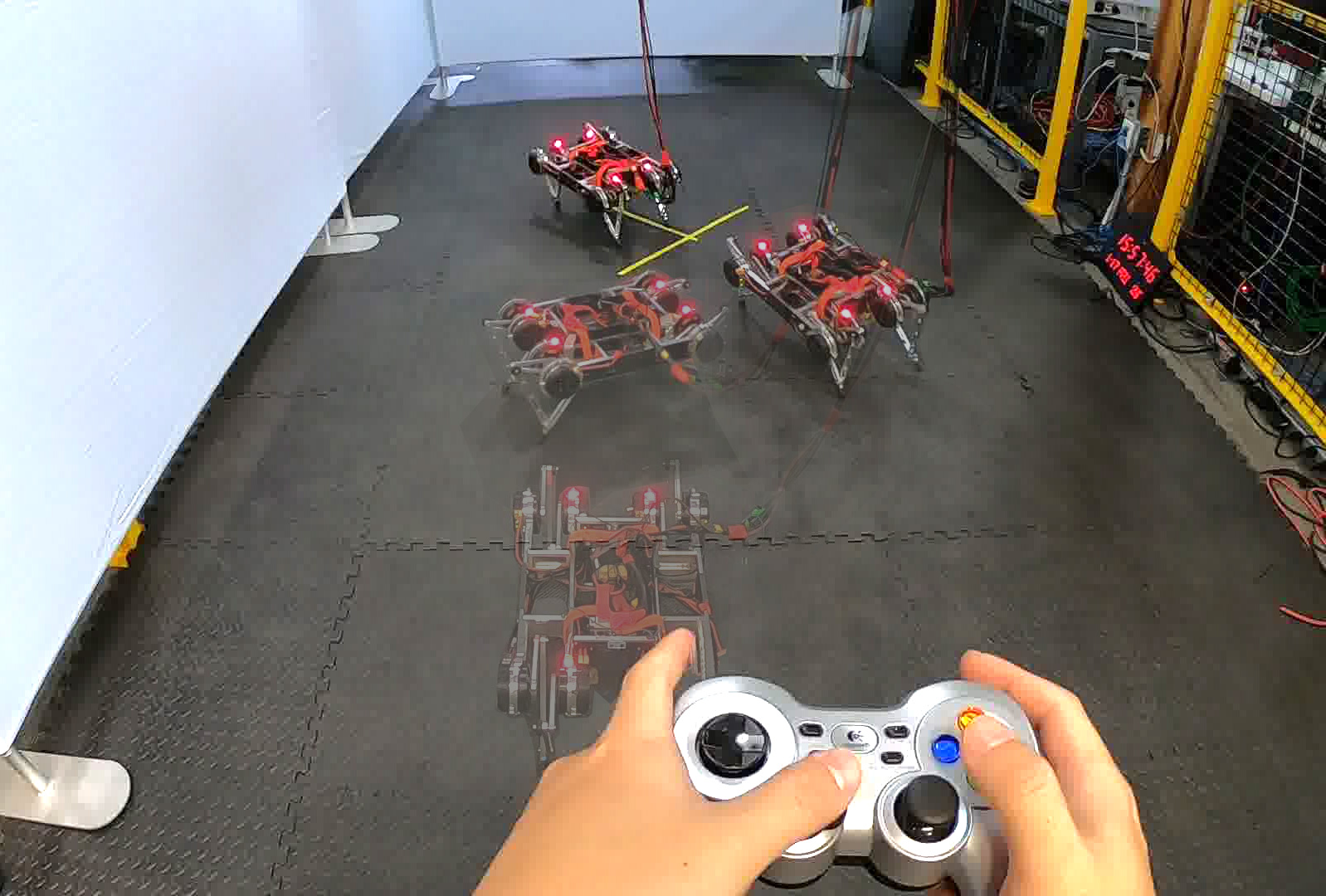} 
  \end{center}
\end{wrapfigure}
We also test our system in the four-task configuration, which includes walking forward, walking backward, turning left, and turning right. 
The learning curve is plotted on Figure~\ref{fig:learning_curves} (2nd). Our system uses in total $320$k samples for all four tasks.
While forward and backward policies are similar to the previous experiment, we obtain effective turning policies that can finish a complete $360^{\circ}$ turn in-place within ten seconds.
Note that turning in-place with the planar leg structure of Minitaur would be difficult to manually-design, but our system can discover it automatically.
We integrated these learned policies with a remote controller, which enables us to switch the gaits and navigate the environment in real-time.

\subsection{Learning to Walk on Varied Surfaces}

\begin{figure*}[tb]
    \centering
    \setlength{\tabcolsep}{1pt}
    \renewcommand{\arraystretch}{0.7}
    \begin{tabular}{c c c c c}
    \includegraphics[width=0.195\textwidth]{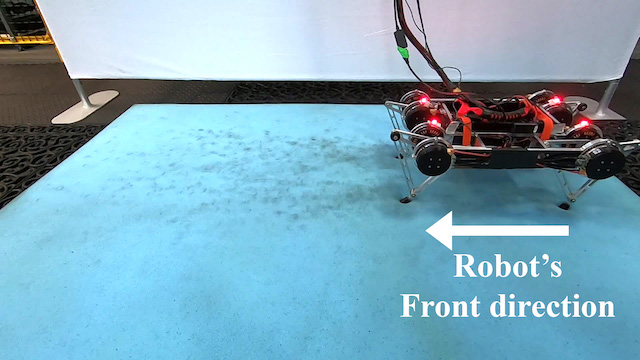} &
    \includegraphics[width=0.195\textwidth]{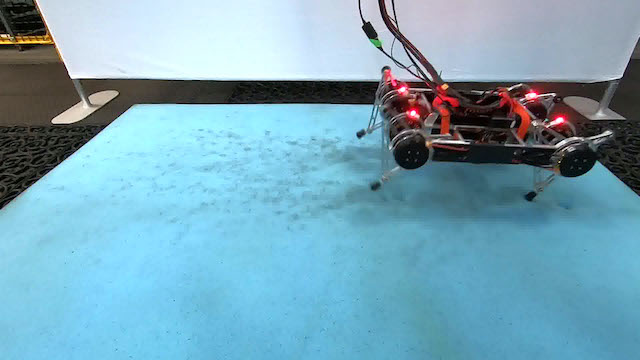} &
    \includegraphics[width=0.195\textwidth]{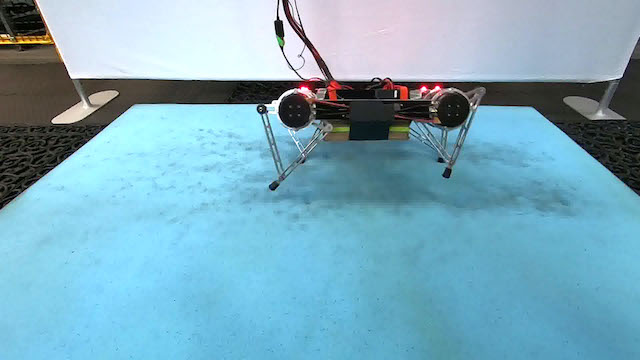} &
    \includegraphics[width=0.195\textwidth]{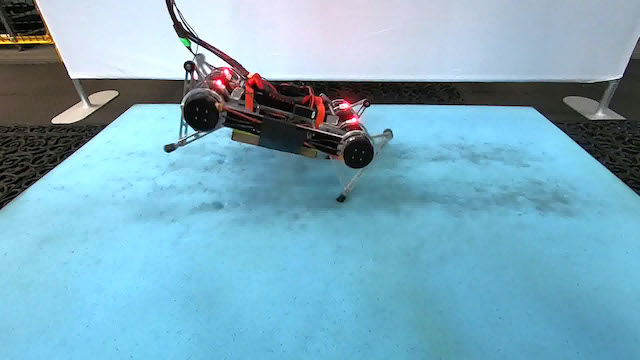} &
    \includegraphics[width=0.195\textwidth]{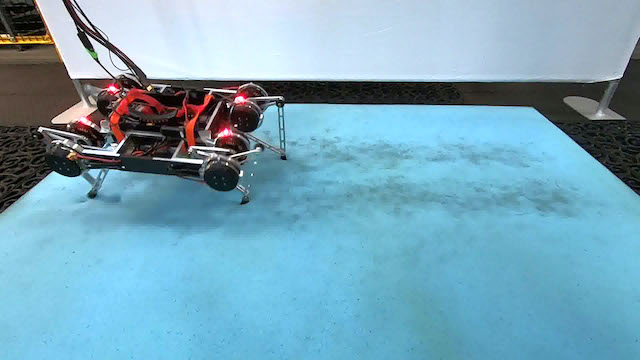} \\
    \includegraphics[width=0.195\textwidth]{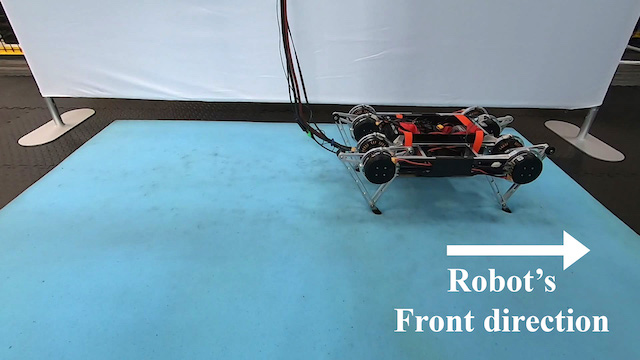} &
    \includegraphics[width=0.195\textwidth]{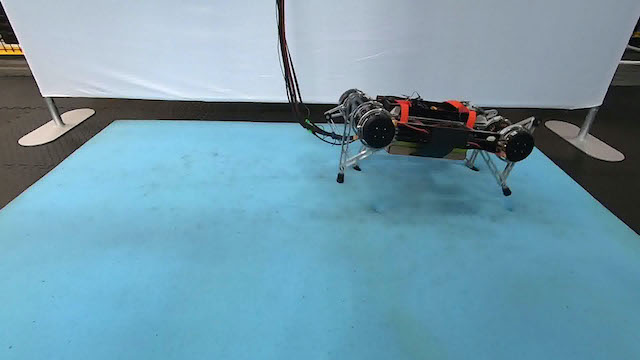} &
    \includegraphics[width=0.195\textwidth]{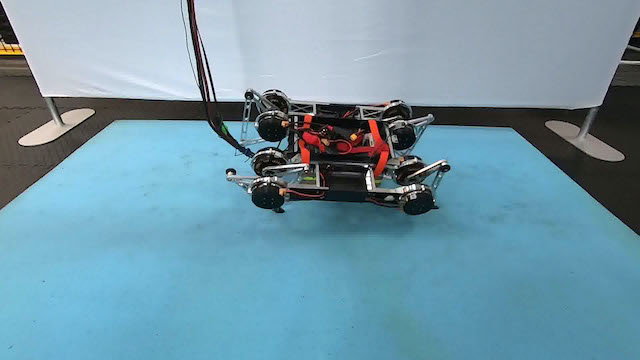} &
    \includegraphics[width=0.195\textwidth]{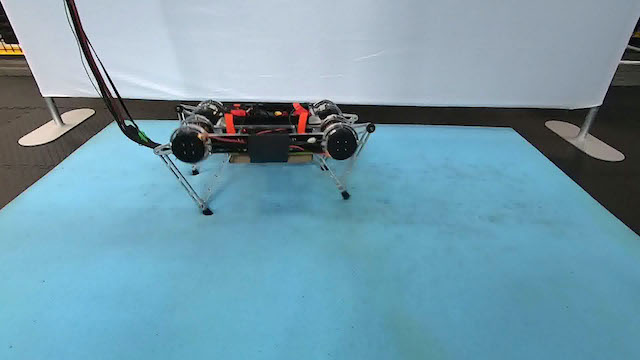} &
    \includegraphics[width=0.195\textwidth]{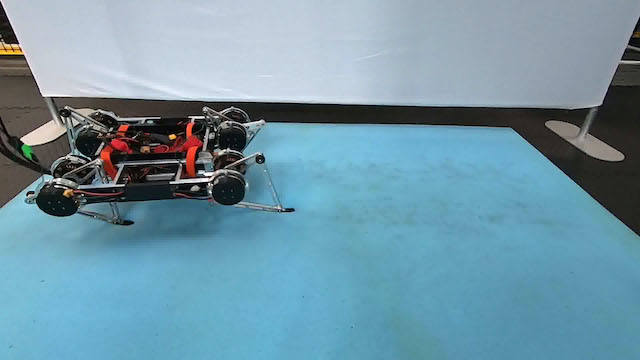} \\
    \includegraphics[width=0.195\textwidth]{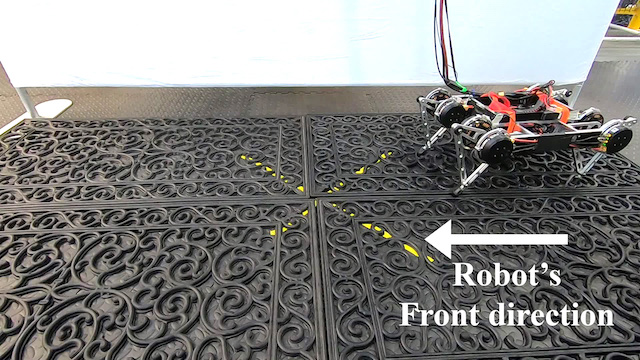} &
    \includegraphics[width=0.195\textwidth]{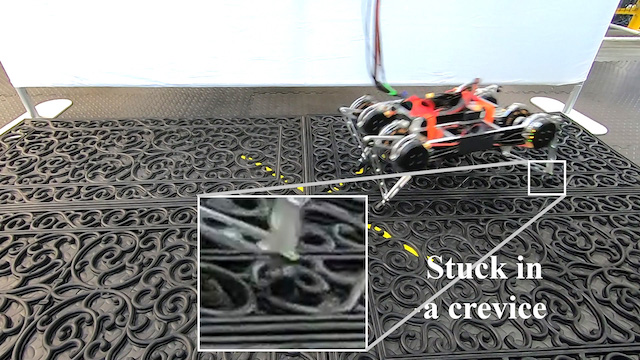} &
    \includegraphics[width=0.195\textwidth]{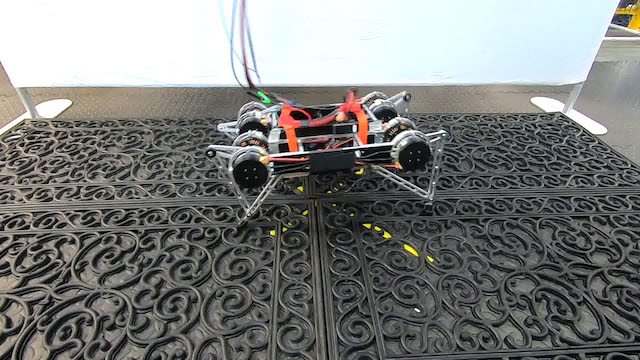} &
    \includegraphics[width=0.195\textwidth]{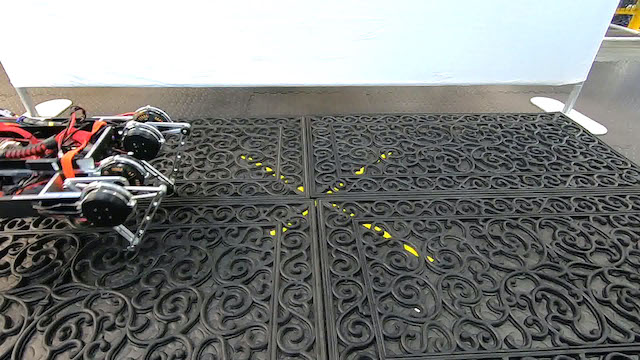} &
    \includegraphics[width=0.195\textwidth]{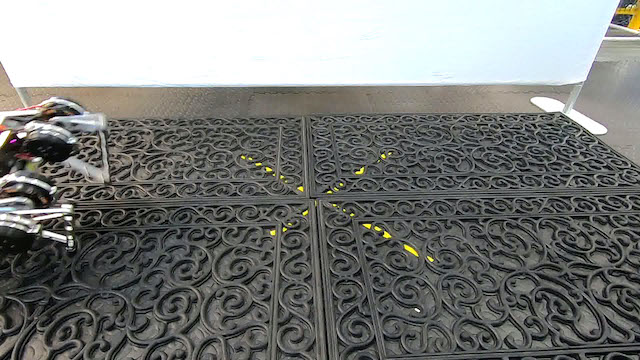} \\
    \includegraphics[width=0.195\textwidth]{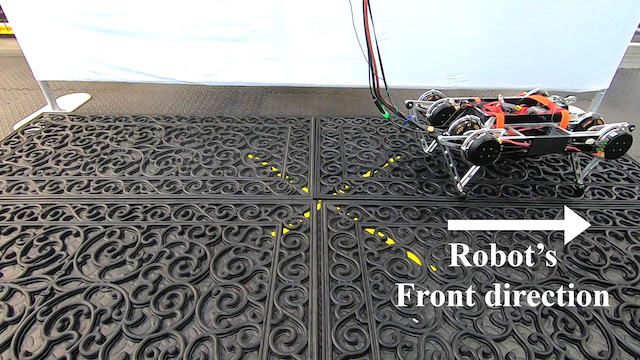} &
    \includegraphics[width=0.195\textwidth]{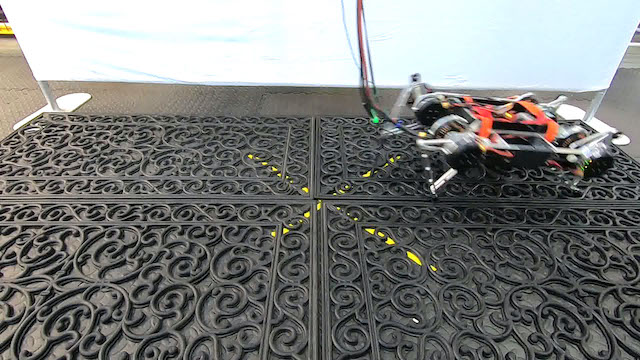} &
    \includegraphics[width=0.195\textwidth]{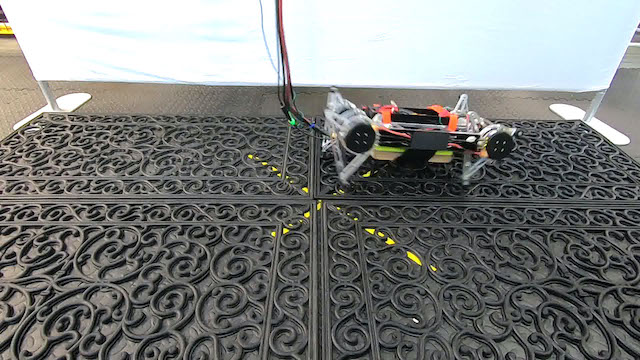} &
    \includegraphics[width=0.195\textwidth]{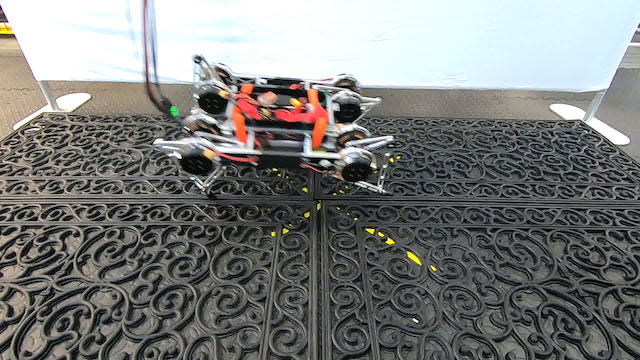} &
    \includegraphics[width=0.195\textwidth]{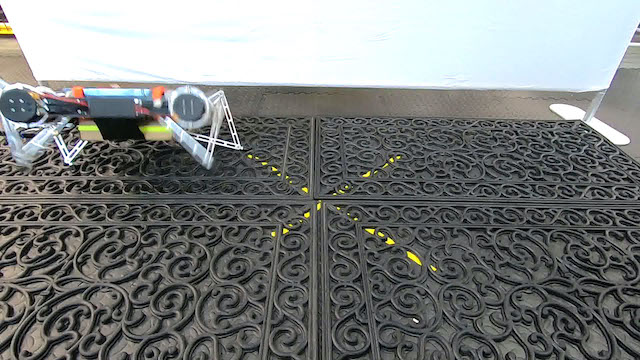} \\
    \end{tabular}
    
    \caption{\small Learned locomotion gaits on the challenging terrains. (\textbf{Row 1}) The learned forward gait {($11$~cm/s)} on the mattress tends to lift the front and back limbs to secure larger ground clearance. (\textbf{Row 2}) The learned backward gait {($13$~cm/s)} on the mattress takes large steps. (\textbf{Row 3}) The learned forward policy {($15$~cm/s)} walks on the doormat while pulling the leg from a crevice. (\textbf{Row 4}) The learned backward gait {($15$~cm/s)} on the doormat resembles an energetic pacing gait.
    }    
	\label{fig:multi_motion}	
\end{figure*}

We also deploy our autonomous learning system to more challenging surfaces: a soft mattress and a doormat with crevices.
The soft mattress is made of gel memory foam ($1.85\times 1.3$~m$^2$, $4$~cm thick).
The doormat is made of rubber ($1.2\times 0.6$~m$^2$, $1.5$~cm thick) with many small details and crevices.
We arranged eight doormats to obtain the $2.4\times 2.4$~m$^2$ workspace.
Both surfaces are \emph{challenging} because walking on these surfaces require special gaits and delicate balance control. {Walking on the soft mattress requires substantially higher foot clearance, while the doormat requires careful movement of the feet to free them when they become caught in crevices. Therefore,}
a policy trained on flat ground cannot walk on either surface. Note that these two surfaces have complex material properties and fine geometric details that are difficult to simulate, further underscoring the importance of real-world training.

Our system successfully learns to walk forward and backward on both surfaces (3rd and 4th subplots in Figure~\ref{fig:learning_curves}).
Training on these surfaces requires more samples than the flat surface. Walking forward and backward required $200$k steps {($5.5$ hours)} for the mattress and $150$k steps {($4.5$ hours)} for the doormat.
In both cases, learning to walk backward was slightly easier than the forward locomotion task.
On the soft mattress, learned policies find gaits with larger ground clearance than the flat ground by tilting the torso more.
The learned forward policy is not homogeneous over time and alternates between pacing, galloping, and pronking gaits, although each segment does not perfectly coincide with the definition of any particular gait (Figure~\ref{fig:multi_motion}, Top).
Locomotion policies on the doormat are more energetic than flat terrain policies in general (Figure~\ref{fig:multi_motion}, Bottom).
We observe that the forward policy often shakes the leg when it is stuck within a crevice -- an effective strategy that is necessary for the doormat.
Although our framework greatly reduces the number of failures, it still requires a handful of manual resets ($20$ to $30$) when training on these challenging surfaces.
We observe that the increased number of manual resets compared to the flat ground is largely due to the reduced size of the workspace. 

\subsection{Analysis}


In this section, we evaluate the two main components of our framework -- multi-task learning and the safety-constrained SAC -- using a large number of training runs in PyBullet simulation~\cite{coumans2013bullet}.
Simulation allows us to easily collect a large quantity of data and compare the statistics of the performance without damaging the robot.
We emphasize that these experiments in simulation are only for analysis, and we do not use any simulation data for the experiments in the previous sections.
For all statistics in this section, we run experiments with five different random seeds.

\begin{figure}
\begin{minipage}{\textwidth}

    \centering
    \setlength{\tabcolsep}{1pt}
    \renewcommand{\arraystretch}{0.7}
    \begin{tabular}{c c c}
    \includegraphics[width=0.33\textwidth]{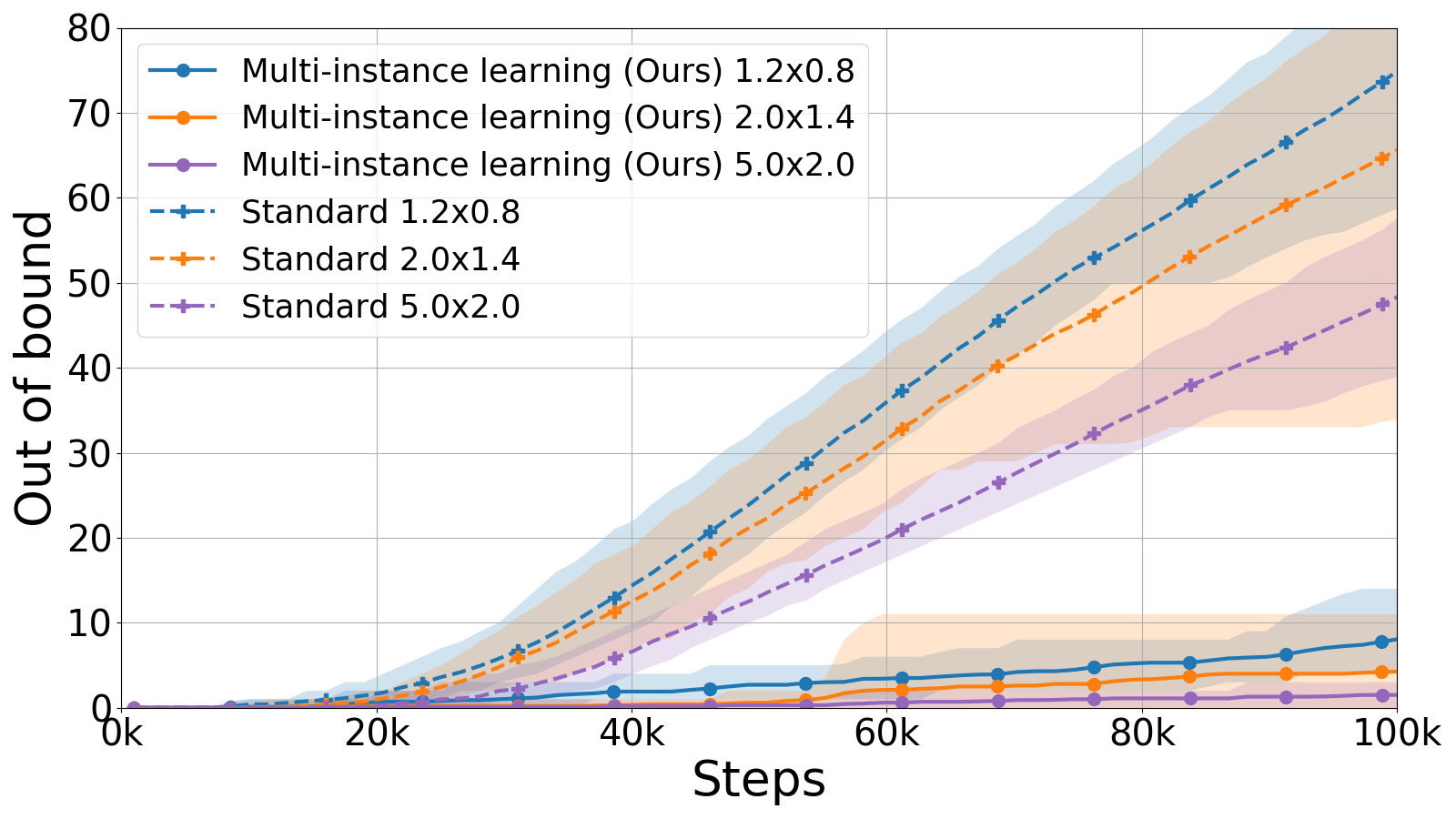}  &  
    \includegraphics[width=0.33\textwidth]{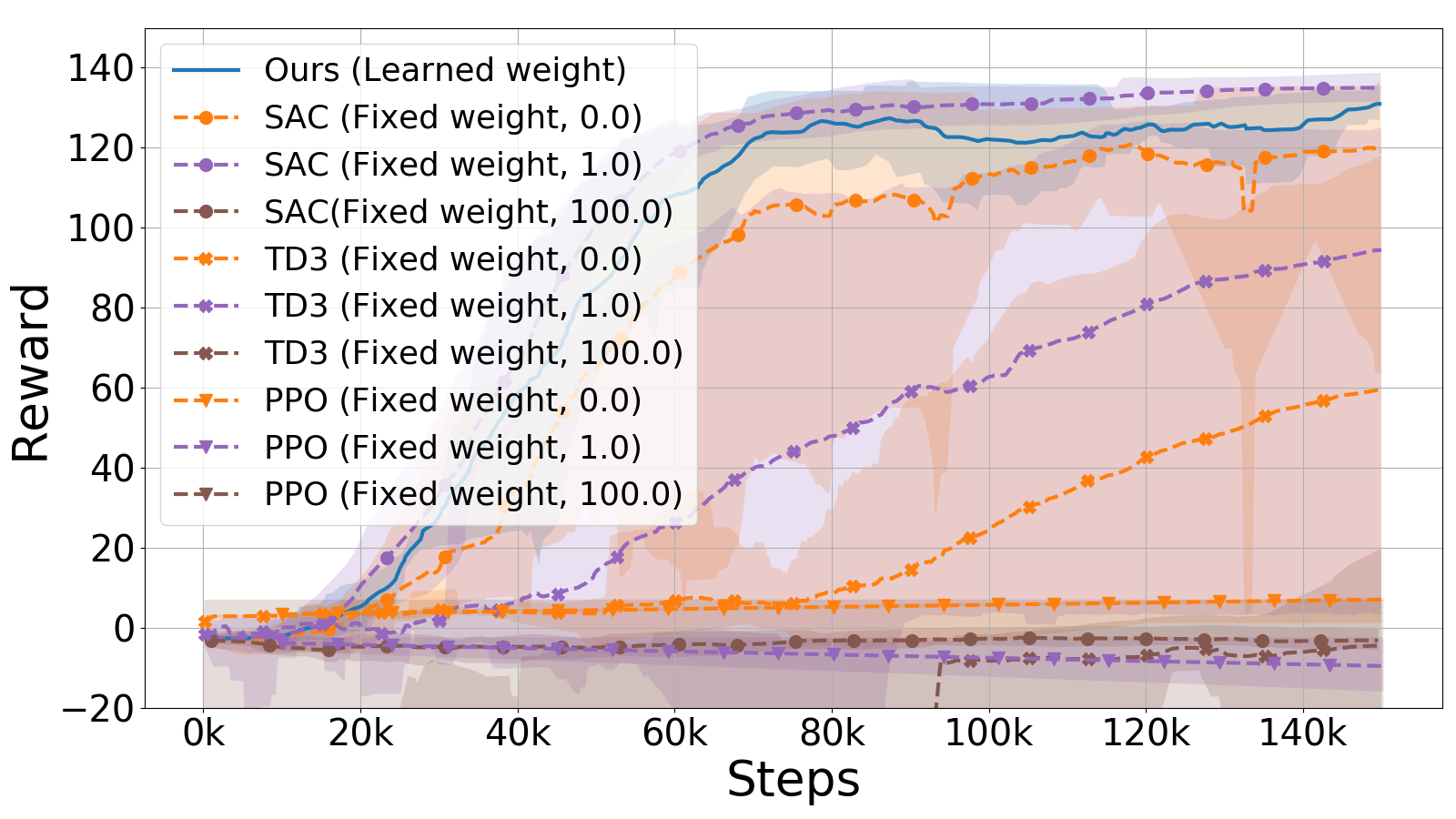} &
    \includegraphics[width=0.33\textwidth]{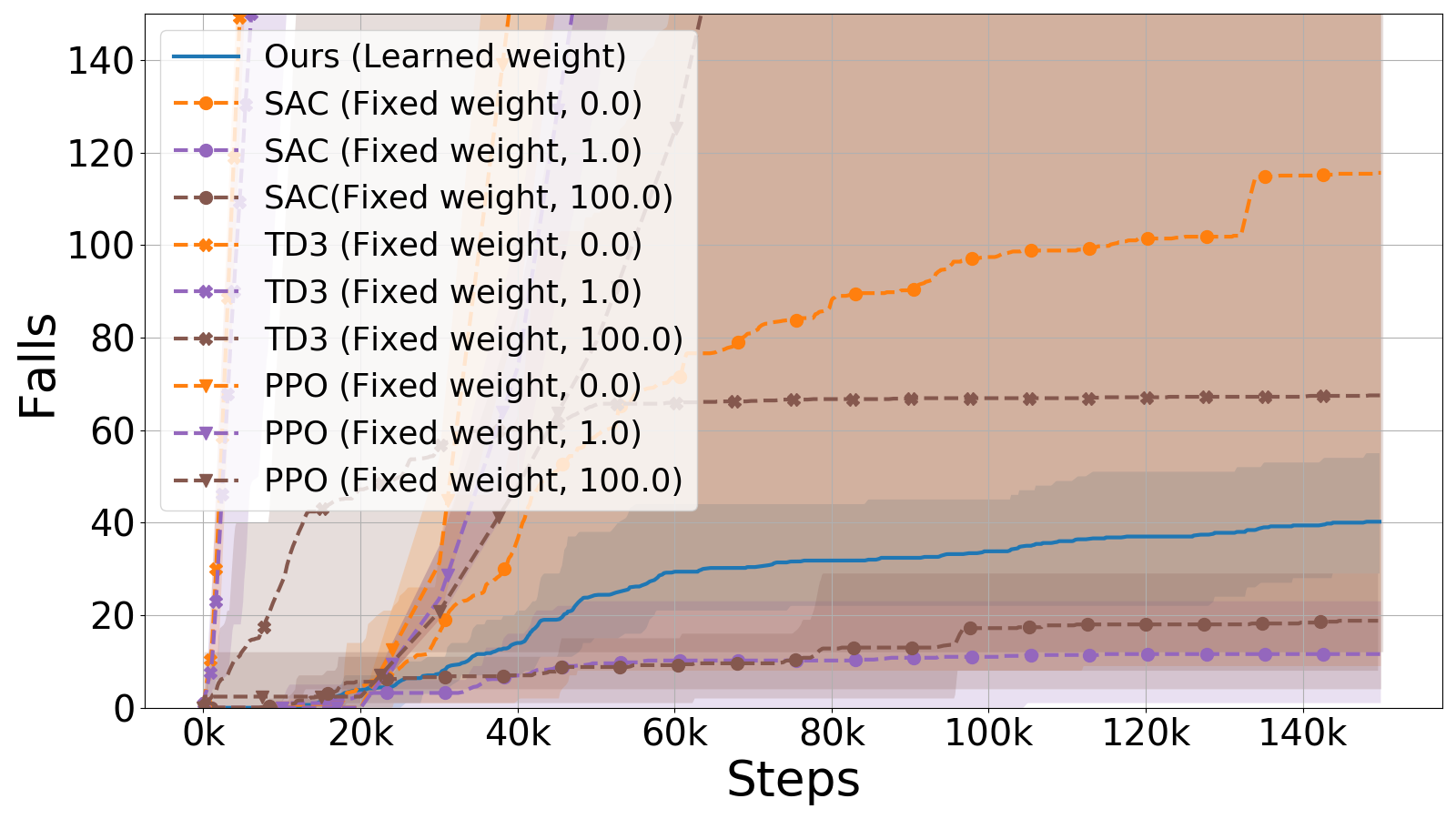} \\
    \end{tabular}
    \caption{\small 
    \textbf{(Left)} The number of out-of-workspace failures with a multi-task learning method (ours) and a standard learning method (baseline).
    For all sizes of workspace, multi-task learning dramatically reduces the number of out-of-workspace failures.
    \textbf{({Middle and Right})}
    Comparison of different learning algorithms and different ways to enforce safety: fix-weighted safety penalties and safety constraints (ours) with a learnable Langrangian multiplier.
    Our method (blue solid curve) reduces the number of falls significantly compared to the baselines without any safety constraint (e.g., SAC with a fixed weight = 0.0).
    Its performance is close to the optimally-tuned safety weight (SAC with a fixed weight = 1.0). Note that our method does not need careful hyperparameter tuning, as it is often infeasible for training on real robots. In the middle plot, the reward of the PPO(w=$100.0$) is near $-150.0$ and excluded from the plot. 
    }
	\label{fig:analysis}
\end{minipage}
\end{figure}

First, we compare the number of out-of-workspace failures for our multi-task learning method and the single-task RL baseline.
Because the number of failures is related to the size of the workspace, we select three sizes ($1.2\times 0.8$m$^2$, $2.0\times 1.4$m$^2$, and $5.0\times 2.0$m$^2$) that are similar to the real-world settings of the mattress, the doormat, and flat ground, respectively.
Figure~\ref{fig:analysis} (Left) shows that our method learns policies with a much smaller number of failures, which is only $5$~\% to $10$~\% of the baseline.
The trend was the same for all three sizes, although smaller workspace requires more manual resets, as expected.
And the number of out-of-workspace failures for multi-task learning methods in a large workspace ($5.0\times 2.0$m$^2$) is also well matched with our empirical results in the real-world experiment.

Figure~\ref{fig:analysis} also shows comparisons of different learning algorithms (PPO~\cite{schulman2017proximal}, TD3~\cite{fujimoto2018addressing}, and SAC~\cite{haarnoja2018learning}), and different ways to enforce safety: using a fixed penalty weight in the reward or a hard constraint (our method). We measure both the reward (Figure~\ref{fig:analysis}, Middle) and the number of falls (Figure~\ref{fig:analysis}, Right) during the learning. First of all, it is clear that our method (blue curve) is one of the most data-efficient, and thus most suited for real-world training. Second, the results in Figure~\ref{fig:analysis} Right indicate that our approach (blue curve) trains a successful policy with approximately $40$ falls, while SAC without safety constraints (Fixed weight = 0.0) falls over 100 times.
However, our method falls more often than the optimally-tuned safety weight (SAC, Fixed weight = 1.0).
In practice, the safety (minimizing failure) and the optimality (maximizing reward) are often contradictory, as observed in the case of an excessive weight of $100.0$ (SAC, Fixed weight = 100.0).
Finding a good trade-off may require extensive hyper-parameter tuning, which is infeasible when training on real robots.
Our method can significantly reduce the number of falls without hype-parameter tuning.

\section{Conclusion and Future Work} 
\label{sec:conclusion}

We presented an autonomous system for learning legged locomotion policies in the real world with minimal human intervention.
We focus on resolving key challenges in automation and safety, which is the bottleneck of robotic learning and is complementary to improving the existing deep RL algorithms.
First, we develop a multi-task learning system that prevents the robot from leaving the training area by scheduling multiple tasks with different target walking directions.
Second, we solve a safety-constrained MDP, which significantly reduces the number of robot falling and breaking during training without additional hyper-parameter tuning.
In our experiments, we show that developing an autonomous learning system is enough to tackle challenging locomotion problems.
It reduced the number of manual resets by more than an order of magnitude compared to the state-of-the-art on-robot training system of \citet{haarnoja2018learning}.
Furthermore, the system allows us to train successful gaits on the challenging surfaces, such as a soft mattress and a doormat with crevices, where the acquisition of an accurate simulation model is expensive.
And we show that we can obtain a complete set of locomotion policies (walking forward, walking backward, turning left, and turning right) from a single learning session.

One requirement of our system is a robust stand-up controller that works for a variety of situations, which is designed manually in the current version.
Although developing an effective stand-up controller was not difficult due to the simple morphology of the Minitaur robot, it would be ideal if we can automatically learn it using RL.
In the near future, we intend to train a recovery policy from the real-world experience using the proposed framework itself by learning locomotion and recovery policies simultaneously~\cite{eysenbach2018leave}.




\clearpage


\bibliography{references}

\clearpage
\section*{Appendix - Experiment Details}

\begin{figure}
    \centering
    \setlength{\tabcolsep}{1pt}
    \renewcommand{\arraystretch}{0.7}
    \begin{tabular}{c c}
    \includegraphics[height=0.42\textwidth]{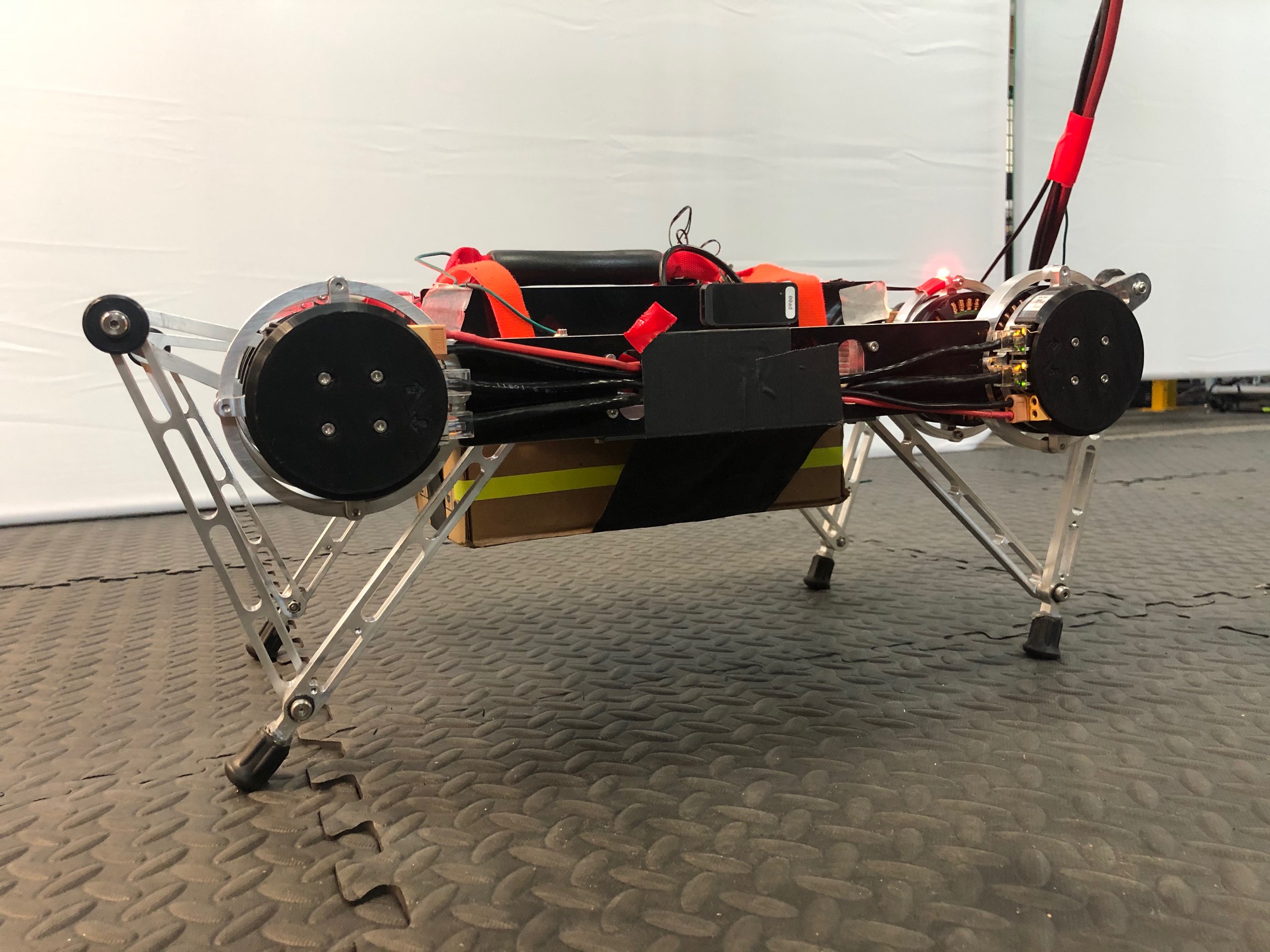} &
    \includegraphics[height=0.42\textwidth]{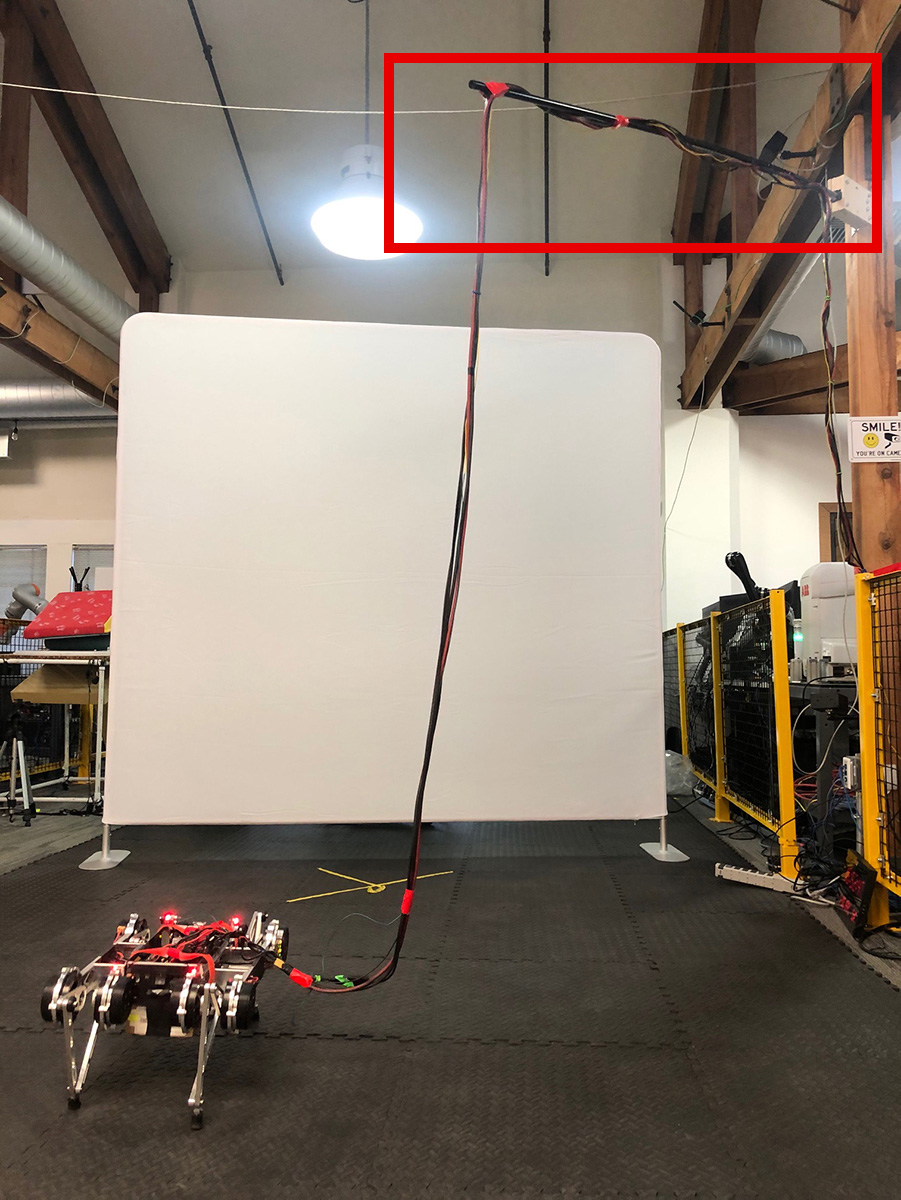} \\
    \end{tabular}  
    \caption{\small 
    Illustration of Hardware.
    (\textbf{Left}) A Minitaur robot with a safety box.
    (\textbf{Right}) A 1-DoF cable management system.
    }
	\label{fig:hardware}
\end{figure}

We test our framework on Minitaur, a quadruped robot (Figure \ref{fig:hardware}, Left), which is approximately $7$~kg with $0.60$m body length.
The robot has eight direct-drive servo motors to move its legs in the sagittal plane.
Each leg is constructed as a four-bar linkage and is not symmetric: the end-effector (one of the bars) is longer and points towards the forward direction.
The robot is equipped with motor encoders that read joint positions and an IMU sensor that measures the torso's orientation and angular velocities in the roll and pitch axes.
In our MDP formulation, the observation consists of motor angles, IMU readings, and previous actions, in the last six time steps.
The robot is directly controlled from a non-realtime Linux workstation (Xeon E5-1650 V4 CPU, 3.5GHz) at about $50$~Hz.
At each time step, we send the action, the target motor angles, to the robot with relatively low PD gains, 0.5 and 0.005. 
While collecting the data from the real-world, we train all the neural networks by taking two gradient steps per control step.
We use Equation~\ref{eq:reward} as the reward function, which is parameterized by task weights $\mathbf{w}$.

We solve the safety-constrained MDP (Equation \ref{eq:constrained_rl} and \ref{eq:safety}) using the off-policy learning algorithm, Soft Actor-Critic (SAC), which has an additional inequality constraint for the policy entropy.
Therefore, we optimize two Lagrangian multipliers, $\alpha$ for the entropy constraint and $\lambda$ for the safety constraint, by applying dual gradient descent to both variables.
We represent the policy and the value functions with fully connected feed-forward neural networks with two hidden-layers, $256$ neurons per layer, and ReLU activation functions.
The network weights are randomly initialized and learned with Adam optimizer with the learning rate of $0.0003$.

Second, we find that the robot often gets tangled by the tethering cables when it walks and turns. We develop a cable management system so that all the tethering cables, including power, communication, and motion capture, are hung above the robot (Figure \ref{fig:hardware}, Right). We wire the cables through a $1.2$~m rod that is mounted at the height of $2.5$~m and adjust the cable’s length to maintain the proper slackness.
Since our workspace has an elongated shape ($5.0$m by $2.0$m), we connect one end of the rod to a hinge joint at the midpoint of the long side of the workspace, which allows the other end of the rod to follow the robot passively.

Third, to reduce the wear-and-tear of the motors due to the jerky random exploration of the RL algorithm, we post-process the action commands with a first-order low-pass Butterworth filter with a cutoff frequency of $5$~Hz. 


\end{document}